\documentclass{article} 
\usepackage{iclr2026_conference,times}

\usepackage{amsmath,amsfonts,bm}

\def\eqref#1{equation~\ref{#1}}

\def\1{\bm{1}}

\DeclareMathAlphabet{\mathsfit}{\encodingdefault}{\sfdefault}{m}{sl}
\SetMathAlphabet{\mathsfit}{bold}{\encodingdefault}{\sfdefault}{bx}{n}

\usepackage{hyperref}
\usepackage{url}
\usepackage{amsmath}
\usepackage{amssymb}
\usepackage[ruled,vlined]{algorithm2e}
\usepackage{graphicx}
\usepackage{xcolor}
\usepackage{listings}
\usepackage{pgfplots}
\pgfplotsset{compat=1.17}

\usepackage{booktabs}
\usepackage{tabularx}
\usepackage{array}
\usepackage[table]{xcolor}

\usepackage{pgfplots}
\pgfplotsset{compat=1.18}

\usepackage{longtable} 
\usepackage{array}     
\usepackage{booktabs}  

\newcommand{\ours}{\textsc{Law\&Order}\xspace}

\title{Law and Order: Tax Law Autoformalization}

\author{\textbf{Sophia S. Han}$^{1, 2}$ \thanks{Please direct correspondence to sophiahsm6@gmail.com.} \quad
\textbf{Yoshiki Takashima}$^{3}$ \thanks{Part of this project was done when Yoshiki Takashima was a Postdoctoral Scholar at Yale Law School.} \quad
\textbf{Anjiang Wei}$^{1}$ \\  \textbf{Zhaoyu Li}$^{4}$ \quad \textbf{Michael Genesereth}$^{1, 2}$\\ 
$^1$Stanford University \quad $^2$Stanford Law School \quad$^3$Pramaana Labs \quad $^4$University of Toronto \\
}

\iclrfinalcopy 
    
\begin{document}

\lstdefinelanguage{lean4}{
  morekeywords={def,theorem,fun,by,match,with,if,then,else,let},
  morekeywords=[2]{native_decide,absurd,congrArg,rfl},
  sensitive=true,
  morecomment=[l]{--},
  morestring=[b]",
  literate={≠}{{$\neq$}}1 {=>}{{$\Rightarrow$}}2,
}
\lstdefinestyle{leanfig}{
  language=lean4,
  basicstyle=\ttfamily\scriptsize,
  keywordstyle=\bfseries\color{blue!60!black},
  keywordstyle=[2]\color{violet!80!black},
  commentstyle=\itshape\color{gray},
  stringstyle=\color{green!40!black},
  columns=fullflexible, keepspaces=true, showstringspaces=false,
  frame=single, rulecolor=\color{black!25},
  aboveskip=0pt, belowskip=0pt,
}

\maketitle

\begin{abstract}

Legal systems are increasingly implemented through software, yet scalable methods for translating legal texts into accurate symbolic representations remain underdeveloped. We study this problem through tax law, where forms and filing instructions define large computational structures involving arithmetic, branching, recursion, and tabular reasoning. We propose \ours, a neuro-symbolic framework for automatically formalizing tax forms and instructions into executable symbolic programs. Our approach establishes two forms of correspondence between law and logic: structural correspondence, which aligns legal and symbolic components such as cells and schedules, and denotational correspondence, which requires symbolic components to implement the computations specified by their legal counterparts. We combine large language model synthesis with cell-level verification and iterative localized error repair using human-written OpenTaxSolver tax returns. We then evaluate the resulting formalizations on independently authored, held-out TaxCalcBench returns, that are never exposed during generation or repair. Although the most advanced LLM achieves only 66\% accuracy, \ours achieves 100\% cell-level and form-level accuracy on 51 held-out returns, demonstrating the effectiveness of combining LLM-based synthesis with symbolic verification for scalable and verifiable large-scale legal autoformalization compared with using an LLM alone.

\end{abstract}
\section{Introduction}
\vspace{-8pt}
Modern societies increasingly rely on software to operationalize law
\citep{delagarza2020automated,lessig_code_2009}. Tax systems, regulatory
frameworks, and other rule-based systems translate legal requirements written
in natural language into executable procedures. Yet automating this translation
remains an open problem. Existing auto-formalization methods are concentrated
primarily in mathematics
\citep{autoformal,zhou2024donttrust,yang2026formal}, while formalization in
expert legal domains remains largely manual, expensive, and difficult to scale
\citep{lawsky2026directfile,sara}. For example, the IRS Direct File system
required \$16 million and more than two years to build
\citep{tigta2026directfile}. Beyond applications to relatively small legal
fragments \citep{smt-put-car-on-stand,lawsky2022coding}, scalable automated
legal formalization has yet to be realized.

Tax law provides a challenging setting for studying this problem. Existing
implementations illustrate its scale: OpenTaxSolver
\citep{roberts2026opentaxsolver} contains more than 16,000 lines of C code, while
implementations of individual forms such as Form 1040 can exceed 1,000 lines.
Moreover, legal rules exist at different levels of abstraction. Statutes and other high-level legal sources are
progressively refined through regulations, guidance, forms, and instructions
before ultimately being implemented in software. We focus on IRS tax forms and their accompanying instructions. These documents
occupy a useful intermediate position between legal text and software. Unlike
statutes and regulations, they are written for ordinary taxpayers and express
many legal requirements in comparatively concrete and operational terms. 

Tax forms also provide a natural structure for formalization and verification.
A form consists of cells containing numeric, Boolean, or finite-sort values.
Some cells are taxpayer inputs, while others are computed from preceding cells
according to natural-language instructions. Formalization therefore requires
translating these instructions into executable arithmetic, branching, tabular
operations, and dependencies among cells. Although individual computations are
often simple, errors can propagate through the dependency structure of a large
form, making final-output verification insufficient for identifying where a
formalization diverges from the intended rules.

They are relatively self-contained: annual revisions are consolidated within form and instruction documents rather than dispersed across amendments and cross-references. Although forms and instructions carry less legal authority than statutes or regulations, they serve as the primary interface through which taxpayers operationalize tax law and determine their tax obligations.

\begin{figure}
    \centering
    \includegraphics[width=0.7\linewidth]{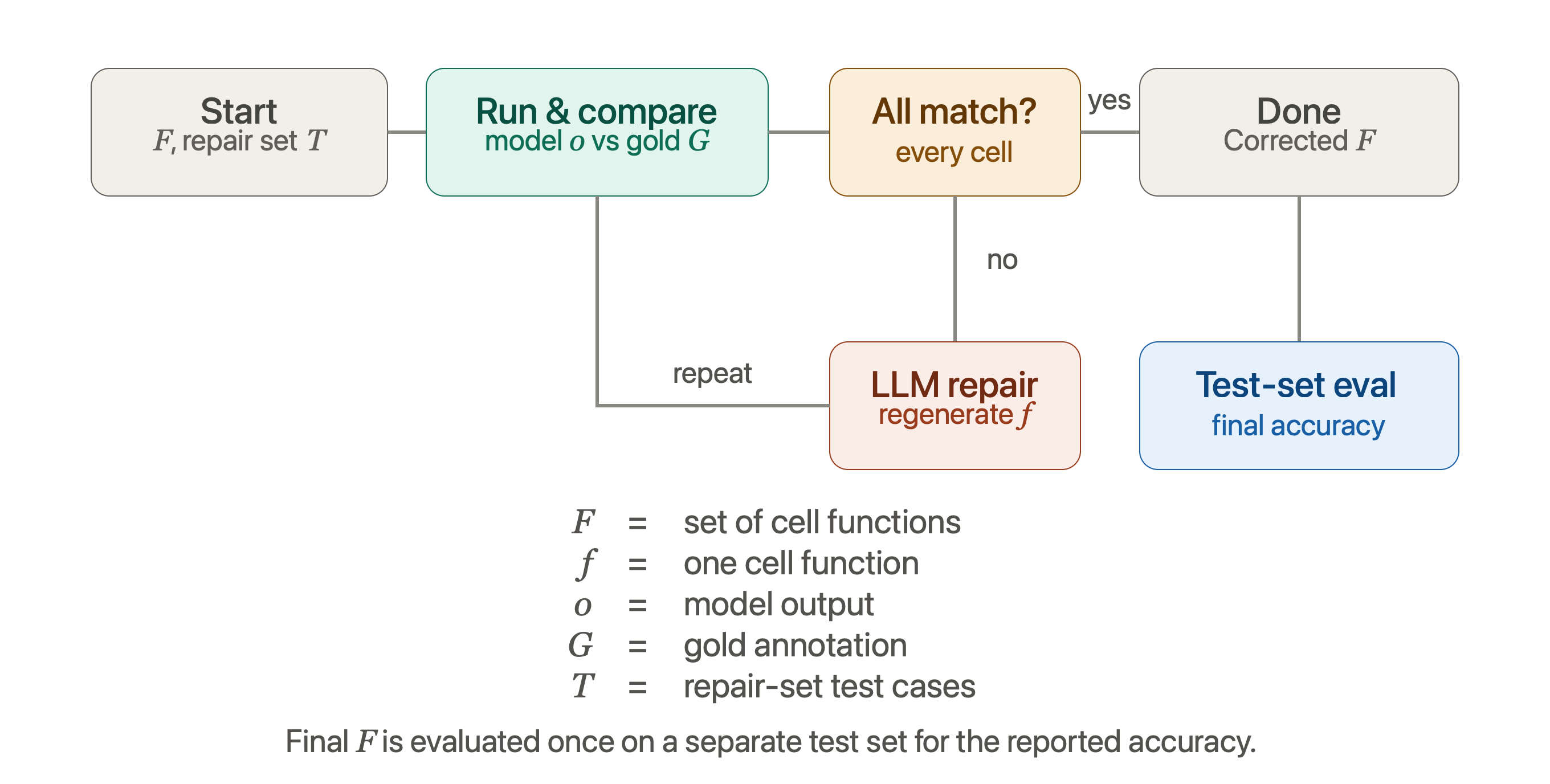}
    \vspace{-10pt}
    \caption{The \ours generation–verification–repair loop. Starting from an initial set of cell functions F and a repair-set of test cases T, the system repeatedly executes F on T and compares each cell's model output o against its gold annotation G. Any cell function f whose output disagrees with G is regenerated by an LLM using that cell's natural-language instruction, and the cycle repeats until every cell matches across all repair-set test cases. The resulting corrected F is then evaluated once on a held-out test set — never seen during synthesis or repair — to produce the final reported accuracy.}
    \label{fig:method_flowchart}
\vspace{-10pt}
\end{figure}

\begin{figure}[t]
\centering

\begin{minipage}{0.96\linewidth}
\centering
\small
\textbf{Lean program: $>10{,}000$ lines}
\vspace{0.3em}

\begin{lstlisting}[
    language=ML,
    basicstyle=\scriptsize\ttfamily,
    xleftmargin=0.5em,
    frame=single,
    framesep=5pt,
    breaklines=true,
    breakatwhitespace=true,
    columns=fullflexible,
    keepspaces=true,
    showstringspaces=false,
    aboveskip=0.4em,
    belowskip=0.2em
]
    /-- 2025 SALT cap (post-OBBBA): base cap $40,000 ($20,000 MFS), reduced by 30% of 
    MAGI (~AGI) over $500,000 ($250,000 MFS), but not below the $10,000 ($5,000 MFS) floor. -/

def cell_072_a5e_limited_state_local_taxes (i : Inputs) : Float :=
  let baseCap   := if i.isMarriedSeparate then 20000.0 else 40000.0
  let threshold := if i.isMarriedSeparate then 250000.0 else 500000.0
  let floorCap  := if i.isMarriedSeparate then 5000.0 else 10000.0
  let reduced   := baseCap - 0.30 * notLessThanZero (i.agi - threshold)
  let cap       := if reduced < floorCap then floorCap else reduced
  smallerOf (cell_071_a5d_state_local_taxes i) cap

def cell_075_a10_total_interest (i : Inputs) : Float :=
  cell_074_a8e_home_mortgage_interest i + inputNum "A9" i

def cell_076_a14_total_gifts_to_charity (i : Inputs) : Float :=
  inputNum "A11" i + inputNum "A12" i + inputNum "A13" i

def cell_077_a17_total_itemized_deductions (i : Inputs) : Float :=
  cell_070_a4_medical_deduction i +
  cell_073_a7_total_taxes i +
  cell_075_a10_total_interest i +
  cell_076_a14_total_gifts_to_charity i +
  inputNum "A15" i +
  inputNum "A16" i

end F1040sa
\end{lstlisting}

\vspace{0.1em}
{\footnotesize\itshape
$\vdots$\quad More than 10,000 lines of reusable executable formalization
\quad$\vdots$}
\end{minipage}

\caption{\textbf{Excerpt from an automatically generated tax program.}
Shown is a small contiguous excerpt of the executable symbolic program
generated from tax forms and filing instructions. The complete formalization
comprises more than 10,000 lines of Lean code, with dependencies spanning
individual cells, worksheets, schedules, and forms. The output of legal
auto-formalization is therefore not merely an isolated formula or short
program, but a large, interconnected executable artifact. This scale
illustrates a central challenge of legal auto-formalization: correctness must be maintained not only for individual rules, but across the computational structure as a whole. It also motivates our focus on compositional synthesis and verification: even when individual rules are relatively compact, their composition produces a program whose correctness cannot be reliably established through generation alone.}
\label{fig:generated-tax-program}
\vspace{-8pt}
\end{figure}

Our central approach is to preserve this structure in the symbolic
representation. We represent computed cells as individual symbolic functions
and organize them according to the sections and schedules of the source form.
This correspondence allows natural-language instructions, symbolic functions,
and execution outputs to be compared at the same granularity. When an output
disagrees with an expected return, the system can localize the discrepancy to
individual cells and revise the corresponding functions. Building on this structure, we present a neuro-symbolic system for tax-law
auto-formalization. The system combines (1) a synthesis module that translates
tax forms and instructions into executable symbolic representations,
(2) a verification module that compares generated programs against expected
cell-level outputs, and (3) an iterative repair procedure that localizes
discrepancies and revises the corresponding symbolic functions. The system
therefore performs test-driven legal formalization, repeatedly generating,
executing, verifying, and repairing the formalization.  Figure~\ref{fig:method_flowchart} illustrates this process. 

Crucially, we use different test cases for development and final evaluation.
During synthesis and repair, the system receives human-written test returns
from OpenTaxSolver~\citep{roberts2026opentaxsolver}. These returns provide
execution feedback and are explicitly available for correcting the generated
formalization. Once this process is complete, the formalization is frozen and
evaluated on a separate set of independently authored returns from TaxCalcBench~\citep{bock2025taxcalcbench}. These evaluation returns
were written by Certified Public Accountants (CPAs) and are never exposed
during generation or correction. The final evaluation therefore measures
whether formalizations developed using one set of human-written cases
generalize to independently written, previously unseen returns.

Our contributions are threefold. First, we formulate large-scale legal
formalization around structurally corresponding representations that align
natural-language rules with executable units. Second, we introduce a
neuro-symbolic synthesis, verification, and repair framework that exploits this
structure for fine-grained error localization.  
Third, we demonstrate that resulting formalizations generalize from human-written development cases to
independently authored, CPA-written held-out returns across 26 tax forms. Although the most advanced LLM can only achieve 66\% accuracy, \ours achieves 100\%, showing the effectiveness of our neuro-symbolic system. 


\vspace{-8pt}
\section{Related Work}
\vspace{-8pt}
\subsection{Autoformalization Methods and Verification}
\vspace{-8pt}
Autoformalization has been studied primarily in mathematics, with benchmarks spanning Olympiad problems, undergraduate theorems, Euclidean geometry, and large-scale Lean 4 corpora \citep{azerbayev2023proofnet,murphy2024autoformalizing,gao2025herald,li2024survey,weng2025autoformalization}. Large language models have shown substantial capability in translating natural-language mathematics into formal specifications \citep{wu2022autoformalizationlargelanguagemodels,zheng2022miniff,azerbayev2023proofnet,murphy2024autoformalizing}, while subsequent work has improved performance on larger formalization tasks through decomposition, dependency information, and iterative correction \citep{liu2025rethinking,wang2025aria,min2026divide}. Evaluation, however, remains challenging: syntactic validity and successful compilation do not guarantee that a generated formalization preserves the meaning of its source. Prior work therefore supplements compilation with semantic evaluation or formal equivalence checks \citep{murphy2024autoformalizing,liu2025rethinking}. This distinction is important for legal formalization, where a well-formed executable program may encode a different rule from the underlying legal text.
\vspace{-8pt}
\subsection{Computational Law and Tax-Law Formalization}
\vspace{-8pt}
Computational-law research has long developed executable representations of legal rules. Catala supports the systematic translation of statutory law into executable programs while preserving default-and-exception structure \citep{merigoux2021catala}, and related work frames tax-software construction as legal formalization \citep{lawsky2022coding,lawsky2026direct}; these approaches, however, largely rely on human-authored formalizations. Legal benchmarks such as SARA and DeonticBench evaluate statutory reasoning and executable Prolog generation \citep{sara,dou2026deonticbenchbenchmarkreasoningrules}. Formal methods have also been used for validation: CUTECat generates tests through concolic execution \citep{goutagny2025cutecat}, while \citet{gogani2025taxprep} use LLM agents and metamorphic testing to synthesize and test tax-preparation software. Our approach instead automatically synthesizes formalizations and compares their executions against expected intermediate cell values, providing localized feedback for repairing discrepancies while preserving correct computation.
\vspace{-8pt}
\section{Method}
\vspace{-8pt}
\begin{algorithm}[t]
\caption{Generation, Verification and Correction of Tax-Form Symbolic Logic}
\label{alg:taxform-repair}

\KwIn{
Test cases with cell-by-cell output $T=\{(I^{(1)},G^{(1)}),\ldots,(I^{(n)},G^{(n)})\}$;\\
cell functions $F=\{f_1,\ldots,f_m\}$;\\
instructions $\mathcal{I}=\{\mathcal{I}_1,\ldots,\mathcal{I}_m\}$;\\
dependency structure over $F$
}
\KwOut{Corrected cell functions $F$}

\Repeat{
$\forall k\in\{1,\ldots,n\},\forall j\in\{1,\ldots,m\},
\; o_j^{(k)}=g_j^{(k)}$
}{
    \tcp{Execute computation graph on all test cases}
    \For{$(I^{(k)},G^{(k)})\in T$}{
        \For{$j\gets 1$ \KwTo $m$}{
            $x_j^{(k)} \gets$ inputs required by $f_j$\;
            $o_j^{(k)} \gets f_j(x_j^{(k)})$\;
        }
    }

    \tcp{Repair functions that disagree with NL reasoning}
    \For{$j\gets 1$ \KwTo $m$}{
        \For{$k\gets 1$ \KwTo $n$}{
          \If{$o_j^{(k)} \neq G^{(k)}_{j}$}{
                $f_j \gets \mathrm{LLMCorrect}\bigl(f_j, \{x_j^{(k)}\}_{k=1}^n,\mathcal{I}_j\bigr)$\;
            }
        }
    }
}

\Return{$F$}

\end{algorithm}

Our approach transforms tax forms and their accompanying instructions into an
executable symbolic representation assisted tax return examples. The
pipeline consists of three stages: (1) form structure extraction, (2) function
signature generation, (3) instruction-grounded program synthesis and 4) repair set-assisted program verification and correction. 

The objective of our system is to synthesize a symbolic program that exactly
reproduces the computations specified by a tax form and its accompanying
instructions. 
\vspace{-8pt}
\subsection{Form structure extraction}
\vspace{-8pt}
Given a tax form, we first use a regex-based algorithm for extracting all computationally relevant cells and
their associated definitions from the PDF file. Each cell and its definition corresponds to a function in the
resulting symbolic representation. 

To evaluate the reliability of this stage, we manually audited the extracted
representations across a diverse collection of federal and state tax forms,
including U.S. Form 1040 and its schedules, as well as tax forms from Arizona,
California, Massachusetts, Michigan, New Jersey, New York, North Carolina,
Ohio, Oregon, Pennsylvania, and Virginia. Across all audited forms, the
extracted cell structures matched the source documents without observed errors.
\vspace{-8pt}
\subsection{Function signature generation}
\vspace{-8pt}
For each extracted cell, we generate an empty function specification
that represents the computation required to determine the value of that cell. At
this stage, only the function interface is generated; the implementation logic
is intentionally omitted. Let (P) denote the synthesized program and let
($T={t_1,\ldots,t_n}$) denote a collection of benchmark tax returns from our repair set. Each repair set example $t_j$ contains a complete set of input
values and expected outputs. We evaluate the synthesized functions by executing
(F) on the repair set example inputs $I_j$ and comparing the resulting cell values against the repair set example outputs $G_j$. We then employ an iterative synthesis–verification loop for symbolic logic generation. 


Formally, each cell ($c_i : \mathbb{N} | \mathbb{R} | \{\mathrm{true}, \mathrm{false}\}$) is associated with a function $f_i : X_i \rightarrow C_i$
where ($X_i$) denotes the set of input variables referenced by the tax form. The resulting collection of function signatures defines the computational skeleton of the tax form while remaining agnostic to the underlying implementation language. The representation can be instantiated in any formal language capable of expressing executable symbolic logic, including Epilogue, C++, Lean, Python, or domain-specific languages. 

As with the extraction stage, we manually verified the correctness of generated function signatures across all audited tax forms and observed no errors.
\vspace{-8pt}
\subsection{Instruction-Grounded Program Synthesis}
\vspace{-8pt}
Tax forms are accompanied by official instructions that specify how each cell should be computed \footnote{https://www.irs.gov/pub/irs-pdf/i1040gi.pdf}. We leverage these instructions as the authoritative semantic source for program generation as they are specifically drafted for ordinary tax-payers, logically well-structured, and require substantially less legal interpretation than statues and Code of Federal Regulations. 

During this step, an LLM receives (1) the tax form, (2) the corresponding instruction text, and (3) the function signatures. The model synthesizes executable symbolic logic for each function signature that implements the computation described in the instructions. Conceptually, this stage performs a translation from natural-language legal rules into formal executable specifications. The synthesized logic may use any formal language that preserves the tax instructions' semantics and supports deterministic execution.

\vspace{-8pt}
\subsection{Repair set-assisted verification and correction}
\vspace{-8pt}
Recent work has increasingly emphasized that syntactic validity alone is
insufficient for evaluating autoformalization. \citet{liu2025rethinking} introduce
an equivalence-based metric for measuring faithful formalization. 
 We introduce two complementary forms of correspondence for law and software. \emph{Structural
correspondence} asks whether components of the legal artifact---such as cells,
schedules, and dependencies---have corresponding components in the symbolic
representation. \emph{Denotational correspondence} asks whether those components
produce the prescribed behavior. Tax forms provide an unusually useful setting
for the latter because benchmark returns expose intermediate cell values.
Discrepancies can therefore be localized to individual symbolic functions and
used to drive targeted repair. Agreement on a finite benchmark establishes
cell-level behavioral agreement on the repair set examples rather than universal
semantic equivalence over all possible taxpayer inputs.

To validate the synthesized program, we compare its outputs—specifically, the cell values computed by each generated function—against an independently generated repair set of tax returns. Each example in the repair set consists of a set of input variables together with the expected values for all cells in the form. Validation proceeds sequentially, comparing cell values step by step. When a discrepancy is identified, the function corresponding to the affected cell is regenerated by prompting the LLM with the relevant tax form and instruction context to correct the implementation. The validation process is then repeated. This iterative refinement cycle continues until the outputs produced through symbolic execution match the repair set examples for every cell in the form. 

Because tax forms define a deterministic computation graph, agreement at every intermediate cell implies agreement for the repair-set examples on the final tax outputs. Consequently, the final synthesized program is accepted only when it reproduces the complete computation trace without any inconsistencies.

\vspace{-8pt}
\section{Experiments}
\vspace{-8pt}
\subsection{Experimental setup}
\vspace{-8pt}

\begin{table}[t]
\centering
\small
\setlength{\tabcolsep}{2.5pt}
\caption{Comparison of model performance on Tax Year (TY) 2025 tax-return
computation, with and without symbolic formalization. For each model, we
report baselines without formalization followed by direct translation,
self-debugging with execution feedback, and \ours with localized execution
feedback. WS stands for Web Search. Best results within each model are shown in bold.}
\label{tab:ty2025}

\resizebox{!}{0.25\textheight}{%
\begin{tabular}{
    >{\raggedright\arraybackslash}p{3cm}
    >{\raggedright\arraybackslash}p{3.1cm}
    >{\centering\arraybackslash}p{1.3cm}
    >{\centering\arraybackslash}p{1.3cm}
    >{\centering\arraybackslash}p{1.3cm}
    >{\centering\arraybackslash}p{1.3cm}
}
\toprule
\textbf{Model} &
\textbf{Method} &
\textbf{Correct returns (strict)} &
\textbf{Correct returns (lenient)} &
\textbf{Correct (by cell)} &
\textbf{Correct (by cell, lenient)} \\
\midrule

GPT-6
& w/o formalization & 38.0\% & 54.0\% & 83.2\% & 88.0\% \\
& w/o formalization + WS & 66.0\% & 78.0\% & 88.6\% & 92.0\% \\
& Direct translation & 95.0\% & 95.0\% & 99.3\% & 99.3\% \\
& Self-debugging & \textbf{100.0\%} & \textbf{100.0\%} &
  \textbf{100.0\%} & \textbf{100.0\%} \\
& \ours & \textbf{100.0\%} & \textbf{100.0\%} &
  \textbf{100.0\%} & \textbf{100.0\%} \\
\midrule

GPT-5.5 (Medium)
& Direct translation & 17.5\% & 17.5\% & 73.3\% & 73.5\% \\
& Self-debugging & 37.5\% & 37.5\% & 75.0\% & 75.4\% \\
& \ours & \textbf{95.0\%} & \textbf{95.0\%} &
  \textbf{99.3\%} & \textbf{99.3\%} \\
\midrule

Claude Opus 5.5
& w/o formalization & 40.0\% & 50.0\% & 83.6\% & 87.3\% \\
& w/o formalization + WS & 64.0\% & 76.0\% & 89.0\% & 92.9\% \\
& Direct translation & 52.5\% & 52.5\% & 89.3\% & 89.3\% \\
& Self-debugging & \textbf{95.0\%} & \textbf{95.0\%} &
  \textbf{99.0\%} & \textbf{99.0\%} \\
& \ours & 85.0\% & 85.0\% & 97.6\% & 97.6\% \\
\midrule

Claude Fable 5.1
& w/o formalization & 40.0\% & 50.0\% & 83.5\% & 86.3\% \\
& w/o formalization + WS & 48.0\% & 64.0\% & 84.4\% & 89.4\% \\
& Direct translation & 55.0\% & 55.0\% & 93.9\% & 94.1\% \\
& Self-debugging & \textbf{92.5\%} & \textbf{92.5\%} &
  \textbf{98.5\%} & \textbf{98.5\%} \\
& \ours & 72.5\% & 72.5\% & 95.2\% & 95.4\% \\
\midrule

Gemini 3.8 Flash
& Direct translation         & 27.5\% & 27.5\% & 78.4\% & 78.4\% \\
& Self-debugging & \textbf{75.0\%} & \textbf{77.5\%} &
  \textbf{96.9\%} & \textbf{97.1\%} \\
& \ours & \textbf{75.0\%} & \textbf{77.5\%} & 96.6\% & 96.8\% \\
\midrule

Kimi K3
& w/o formalization & 6.0\% & 12.0\% & 64.5\% & 68.4\% \\
& Direct translation & 62.5\% & 62.5\% & 90.8\% & 90.8\% \\
& Self-debugging & \textbf{95.0\%} & \textbf{95.0\%} &
  \textbf{98.6\%} & \textbf{98.6\%} \\
& \ours & 85.0\% & 85.0\% & 96.3\% & 96.3\% \\
\midrule

Qwen 3.6 8B
& Direct translation & 0.0\% & 0.0\% & 54.7\% & 55.0\% \\
& Self-debugging & 0.0\% & 0.0\% & 54.8\% & 55.2\% \\
& \ours & 0.0\% & 0.0\% &
  \textbf{55.7\%} & \textbf{55.9\%} \\


\bottomrule
\end{tabular}
}
\vspace{-10pt}
\end{table}


\paragraph{Models.}
For symbolic logic translation and correction, we evaluate seven language models spanning a range of capabilities and scales: proprietary frontier models GPT-6~\citep{openai_gpt6_astra}, Claude Opus 5.5~\citep{anthropic_opus55}, Claude Fable 5.1~\citep{anthropic_fable51}, Gemini Flash 3.8~\citep{google_gemini35}, and Gemini 3.1 Pro Preview~\citep{google_gemini31pro} and open-weight models Kimi K3~\citep{moonshot_kimik3} and Qwen 3.6 8B~\citep{qwen36}.
\vspace{-10pt}
\paragraph{Repair set.}
We use test cases generated from OpenTaxSolver as the repair set during the program-repair stage, where failing test cases are used to diagnose and modify the program. OpenTaxSolver (OTS) is a free,
open-source tax preparation package \citep{roberts2026opentaxsolver}. OTS 2024 contains 24 form programs and the tax solver accounts for 14,114 of those lines, and the 2024 Form 1040 solver is 2,441 lines. OTS 2025 contains 27 form programs and the tax solver has 16,060 lines, the shared routines file is 1,369 lines, and the 2025 Form 1040 solver is 2,625 lines. The three programs added in 2025 are Schedule 1-A (282 lines), Form 2441 (654 lines), and Form 6781 (542 lines). 



\vspace{-10pt}
\paragraph{Test set.}
For evaluating our system, we use the 51 tax-return cases from TaxCalcBench. Each case provides taxpayer inputs, the corresponding expected return output, and a testing harness that verifies whether a model produces the correct result. TaxCalcBench includes both federal and state tax returns and covers a range of complex tax and financial scenarios.
\vspace{-10pt}
\paragraph{Baselines}
We compare five approaches to tax-return computation. Without formalization directly computes tax returns from taxpayer inputs and tax instructions without constructing an executable symbolic representation. Without formalization + Web Search (WS) follows the same approach but additionally allows the model to retrieve relevant information from the Web. Direct translation~\citep{han-etal-2024-folio, olausson2023linc} translates tax forms and instructions into an executable symbolic program in a single pass. Self-debugging~\citep{chen2024teaching} iteratively revises the generated program using execution feedback. Finally, \ours performs targeted program repair using localized execution feedback.

\vspace{-8pt}
\subsection{Accuracy measurement}

\vspace{-10pt}
We define cell-level accuracy as
\begin{equation}
\scriptsize
\text{cell Accuracy}
=
\frac{
\#\text{correctly computed cells}
}{
\#\text{total evaluated cells}
}.
\end{equation}
A cell is considered correct if its value matches the output value for the corresponding test case.
A synthesized form is considered correct if all cells match the benchmark outputs across all evaluation cases. Form-level accuracy is therefore defined as
\begin{equation}
\scriptsize
\text{Return Accuracy}
=
\frac{
\#\{\text{forms with zero incorrect cells}\}
}{
\#\{\text{total forms}\}
}.
\end{equation}
\vspace{-10pt}
\subsection{Main results}
\vspace{-8pt}

\begin{table}[t]
\centering
\caption{Comparison of model performance on Tax Year (TY) 2024 tax-return
computation using symbolic formalization. For each model, we compare direct
translation, self-debugging with execution feedback, and \ours with localized
execution feedback. Best results within each model are shown in bold.}
\label{tab:ty2024}
\small

\resizebox{\textwidth}{!}
{%
\begin{tabular}{
    >{\raggedright\arraybackslash}p{3.5cm}
    >{\raggedright\arraybackslash}p{3.5cm}
    >{\centering\arraybackslash}p{1.7cm}
    >{\centering\arraybackslash}p{1.7cm}
    >{\centering\arraybackslash}p{1.7cm}
    >{\centering\arraybackslash}p{1.7cm}
}
\toprule
\textbf{Model} &
\textbf{Method} &
\textbf{Correct returns (strict)} &
\textbf{Correct returns (lenient)} &
\textbf{Correct (by cell)} &
\textbf{Correct (by cell, lenient)} \\
\midrule

GPT-6
& Direct translation
& \textbf{100.0\%}
& \textbf{100.0\%}
& \textbf{100.0\%}
& \textbf{100.0\%} \\

& Self-debugging
& 86.3\%
& 86.3\%
& 97.5\%
& 97.5\% \\

& \ours
& \textbf{100.0\%}
& \textbf{100.0\%}
& \textbf{100.0\%}
& \textbf{100.0\%} \\
\midrule

GPT-5.5 (Medium)
& Direct translation
& 29.4\%
& 31.4\%
& 91.1\%
& \textbf{91.4\%} \\

& Self-debugging
& \textbf{72.5\%}
& \textbf{72.5\%}
& \textbf{94.4\%}
& \textbf{94.4\%} \\

& \ours
& 35.3\%
& 35.3\%
& 91.4\%
& 91.4\% \\
\midrule

Claude Opus 5.5
& Direct translation
& \textbf{92.2\%}
& \textbf{92.2\%}
& \textbf{98.3\%}
& \textbf{98.3\%} \\

& Self-debugging
& 76.5\%
& 76.5\%
& 96.5\%
& 96.5\% \\

& \ours
& \textbf{92.2\%}
& \textbf{92.2\%}
& \textbf{98.3\%}
& \textbf{98.3\%} \\

\bottomrule
\end{tabular}%
}

\vspace{-10pt}
\end{table}

Table~\ref{tab:ty2025} shows that \ours’s primary strength is its ability to substantially improve weaker initial formalizations through localized execution feedback, with the clearest gains appearing for GPT-5.5 (Medium). For this model, LAW\&ORDER increases strict return-level accuracy from only 17.5\% under direct translation and 37.5\% under self-debugging to 95.0\%, while increasing cell-level accuracy from 73.3\% and 75.0\% to 99.3\%, respectively. This large gap over self-debugging suggests that localized feedback can be particularly effective when global debugging fails to identify or isolate errors. LAW\&ORDER also reaches 100\% return- and cell-level accuracy with GPT-6 and substantially improves over direct translation for several other models. However, LAW\&ORDER does not consistently outperform conventional self-debugging: Claude Opus 5.5, Claude Fable 5.1, and Kimi K3 perform better with self-debugging, while the two approaches are comparable for Gemini 3.8 Flash. We hypothesize that this difference may arise because Claude models have been optimized for global self-debugging and agentic coding workflows, making them particularly effective at repairing programs from execution feedback without explicit localization. Finally, LAW\&ORDER cannot fully compensate for insufficient underlying model capability, as Qwen 3.6 8B remains at 0\% return-level accuracy despite modest improvements at the cell level. Overall, these results suggest that LAW\&ORDER is especially valuable when global self-debugging is ineffective—as illustrated by the large improvement for GPT-5.5—although its advantage is model-dependent.

Importantly, the results of self-debugging and \ours should not be interpreted as the accuracy of a single LLM generation pass. Rather, they reflect the accuracy of the complete synthesis–verification-repair system, which combines program generation, symbolic execution, discrepancy localization, and iterative correction until consistency is achieved. We also show the results of model performance with formalization on Tax Year 2024 in Table~\ref{tab:ty2024}. GPT-5.5 performs relatively poorly with \ours on Tax Year 2024 primarily because it fails to correctly formalize Form 1040 Line 35a, which computes the tax refund. This cell is relatively complex, involving four operations, including a division. Because Line 35a lies on the main computation path, this error affects all returns that result in a refund, substantially reducing return-level accuracy. Returns without a refund are unaffected by this error.

Overall, these results demonstrate the value of explicit symbolic structure in domains such as taxation and law, where correctness depends on adherence to formal rules rather than linguistic plausibility alone. Combining natural-language understanding with executable symbolic representations enables intermediate computations to be inspected, verified, and repaired. By grounding computation in explicit symbolic structures, \ours facilitates fine-grained verification and localized correction of generated formalizations.


\vspace{-8pt}
\subsection{Proving Semantic Equivalence and Divergence}
\vspace{-8pt}

We uniformly sampled 50 cells and compare the implementations generated by Claude Opus 5.5 with their corresponding gold implementations. We use Claude Code to construct formal proofs of equivalence or, when the implementations differ semantically, to identify inputs that demonstrate their divergence. For 43 of the 50 sampled cells, equivalence is successfully proven. For the remaining seven cells, Claude Code identifies counterexamples under which the generated and gold implementations produce different outputs. Manual inspection shows that six of these divergences result from errors in the Opus 5.5 implementations, while one reveals an error in the gold implementation: it assigns a \$400,000 benefit to qualifying widowed taxpayers rather than the \$200,000 specified in the tax instructions. Examples of equivalence proofs and counterexamples are provided in Appendix~\ref{sec:semantic_equivalence}.

\vspace{-8pt}
\section{Analysis}
\vspace{-8pt}
\subsection{Complexity analysis of the test set.}
\vspace{-8pt}
\begin{figure}[t]
\centering

\begin{minipage}[t]{0.48\textwidth}
    \centering
    \includegraphics[width=\textwidth]{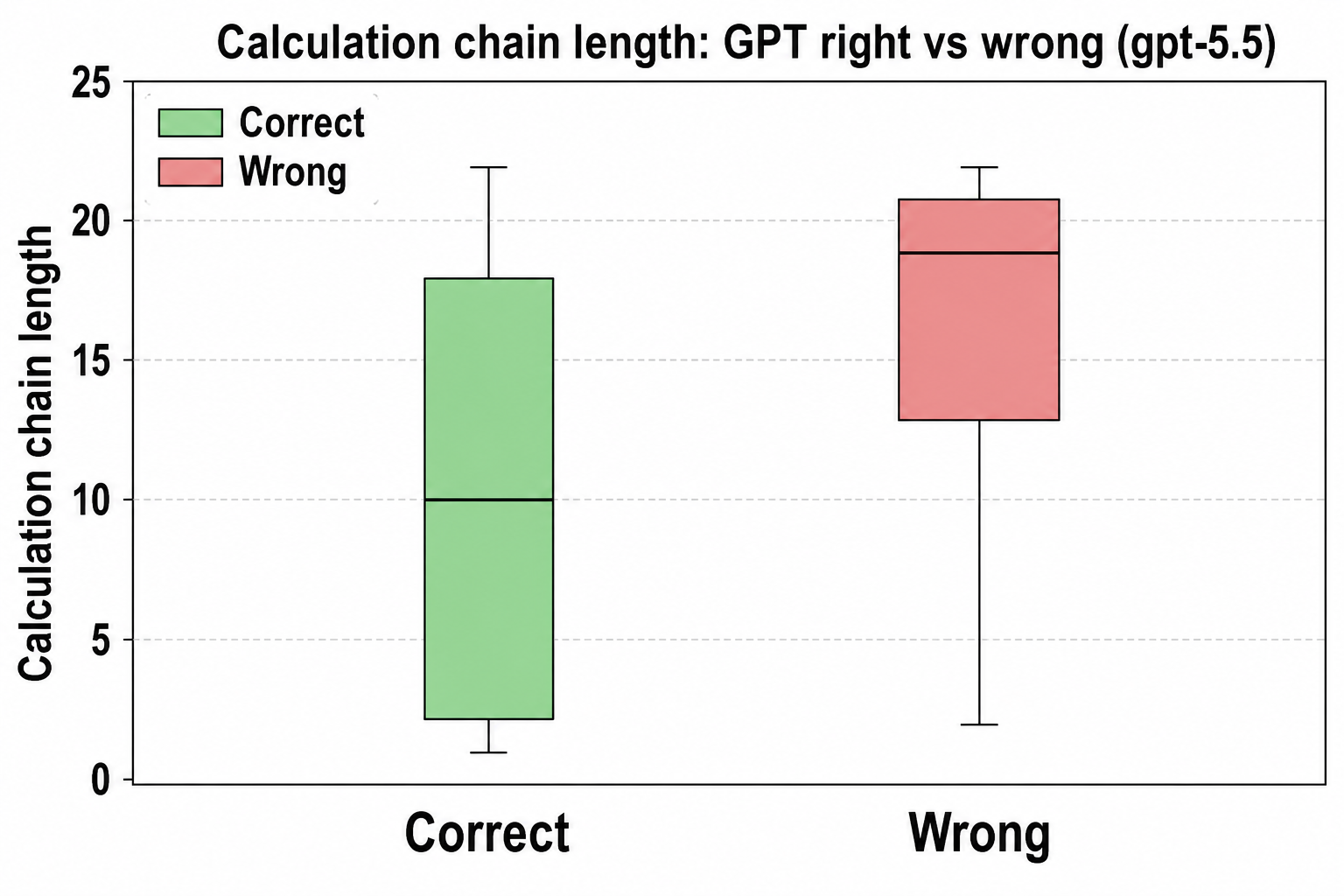}
    \caption{
Distribution of calculation chain lengths for correct and incorrect GPT-5.5 predictions. 
}
    \label{fig:chain_error}
\end{minipage}
\hfill
\begin{minipage}[t]{0.48\textwidth}
\centering
\begin{tikzpicture}
\begin{axis}[
    width=\linewidth,
    height=5cm,
    ymode=log,
    ymin=0.5,
    ymax=2000,
    xmin=-0.6,
    xmax=30.6,
    xlabel={Reasoning depth},
    ylabel={\# of cells (log scale)},
    xtick={0,5,10,15,20,25,30},
    yminorticks=false,
    log ticks with fixed point,
    ymajorgrids=true,
    grid style={gray!25},
    bar width=0.72,
    label style={font=\scriptsize},
    tick label style={font=\scriptsize},
]
\addplot+[
    ybar,
    fill=teal!70!black,
    draw=none,
    forget plot
] coordinates {
    (0,1021) (1,227) (2,119) (3,86) (4,55)
    (5,48) (6,44) (7,35) (8,26) (9,24)
    (10,19) (11,17) (12,14) (13,10) (14,11)
    (15,10) (16,11) (17,14) (18,10) (19,6)
    (20,7) (21,8) (22,4) (23,4) (24,6)
    (25,5) (26,1) (27,1) (28,2) (29,1)
    (30,1)
};
\end{axis}
\end{tikzpicture}

\caption{Histogram of cell count by reasoning depth, pooled across all
26 benchmark forms ($N=1{,}847$). Depth~0 (raw inputs) accounts for
1{,}021 cells; a long tail of multi-step tax logic extends to depth~30.}
\label{fig:cells-by-depth}
\end{minipage}
\vspace{-5pt}
\end{figure}

Figure~\ref{fig:cells-by-depth} shows the distribution of reasoning depth across the 1,847 cells in the 26 benchmark forms. While 1,021 cells correspond to raw inputs at depth 0, the remaining cells span a broad range of reasoning depths, reflecting computations that depend on increasingly long chains of intermediate values. A substantial number of cells require multiple levels of reasoning, with depths of 5, 10, and beyond representing nontrivial dependency chains. The distribution extends to a maximum depth of 30, illustrating that tax-form computation can involve deeply nested dependencies and long sequences of interconnected calculations.
\vspace{-10pt}


\subsection{Accuracy vs length of calculation chains}
\vspace{-8pt}



From Figure~\ref{fig:chain_error}, we show that errors are concentrated among examples with longer calculation chains, indicating that model accuracy decreases as reasoning depth increases. This pattern is consistent with the hypothesis that intermediate mistakes accumulate over multi-step computations, leading to a higher probability of failure on long reasoning trajectories. Similar phenomena have been reported in prior deductive reasoning and math reasoning benchmarks \citep{han-etal-2024-p, miner2025scheherazade} where model performance degrades as the length and complexity of the reasoning chain increase.


    

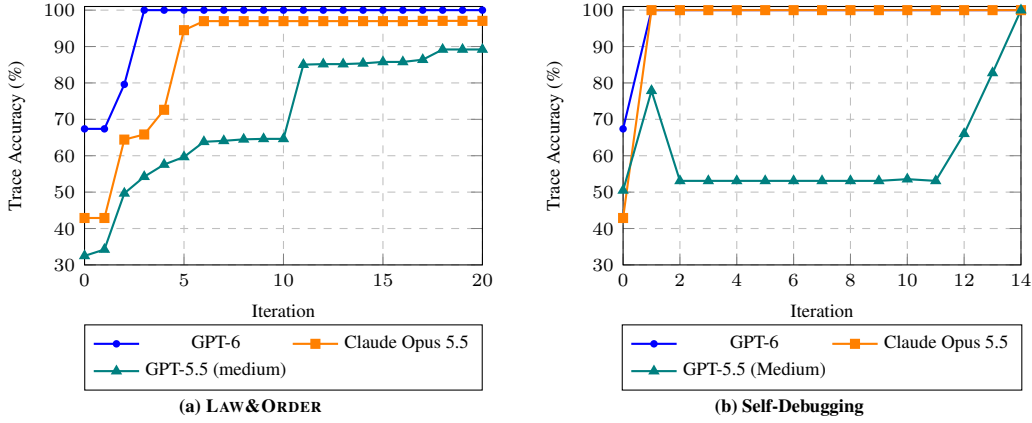
\begin{figure*}[t]
\centering

\begin{minipage}[t]{0.49\textwidth}
\centering
\begin{tikzpicture}
\begin{axis}[
    width=\linewidth,
    height=5cm,
    xlabel={Iteration},
    ylabel={Trace Accuracy (\%)},
    xmin=0, xmax=20,
    ymin=30, ymax=101,
    xtick={0,5,10,15,20},
    ytick={30,40,50,60,70,80,90,100},
    grid=both,
    major grid style={dashed},
    minor grid style={dotted},
    tick label style={font=\scriptsize},
    label style={font=\scriptsize},
    legend style={
        at={(0.5,-0.23)},
        anchor=north,
        legend columns=2,
        font=\scriptsize
    },
]

\addplot[
    thick,
    mark=*,
    mark size=1pt,
    color=blue,
] coordinates {
    (0,67.38)
    (1,67.38)
    (2,79.60)
    (3,100.00)
    (4,100.00)
    (5,100.00)
    (6,100.00)
    (7,100.00)
    (8,100.00)
    (9,100.00)
    (10,100.00)
    (11,100.00)
    (12,100.00)
    (13,100.00)
    (14,100.00)
    (15,100.00)
    (16,100.00)
    (17,100.00)
    (18,100.00)
    (19,100.00)
    (20,100.00)
};
\addlegendentry{GPT-6}

\addplot[
    thick,
    mark=square*,
    mark size=1.5pt,
    color=orange,
] coordinates {
    (0,42.89)
    (1,42.89)
    (2,64.43)
    (3,65.84)
    (4,72.62)
    (5,94.50)
    (6,96.98)
    (7,96.98)
    (8,96.98)
    (9,96.98)
    (10,96.98)
    (11,96.98)
    (12,96.98)
    (13,96.98)
    (14,96.98)
    (15,96.98)
    (16,96.98)
    (17,97.05)
    (18,97.05)
    (19,97.05)
    (20,97.05)
};
\addlegendentry{Claude Opus 5.5}

\addplot[
    thick,
    mark=triangle*,
    mark size=1.7pt,
    color=teal,
] coordinates {
    (0,32.48)
    (1,34.23)
    (2,49.66)
    (3,54.23)
    (4,57.58)
    (5,59.66)
    (6,63.83)
    (7,64.09)
    (8,64.50)
    (9,64.63)
    (10,64.63)
    (11,85.03)
    (12,85.17)
    (13,85.17)
    (14,85.37)
    (15,85.77)
    (16,85.77)
    (17,86.38)
    (18,89.19)
    (19,89.19)
    (20,89.19)
};
\addlegendentry{GPT-5.5 (medium)}

\end{axis}
\end{tikzpicture}

\scriptsize
\textbf{(a) \ours}
\end{minipage}
\hfill
\begin{minipage}[t]{0.49\textwidth}
\centering
\begin{tikzpicture}
\begin{axis}[
    width=\linewidth,
    height=5cm,
    xlabel={Iteration},
    ylabel={Trace Accuracy (\%)},
    xmin=0, xmax=14,
    ymin=30, ymax=101,
    xtick={0,2,4,6,8,10,12,14},
    ytick={30,40,50,60,70,80,90,100},
    grid=both,
    major grid style={dashed},
    minor grid style={dotted},
    tick label style={font=\scriptsize},
    label style={font=\scriptsize},
    legend style={
        at={(0.5,-0.23)},
        anchor=north,
        legend columns=2,
        font=\scriptsize
    },
]

\addplot[
    thick,
    mark=*,
    mark size=1pt,
    color=blue,
] coordinates {
    (0,67.38)
    (1,100.00)
    (2,100.00)
    (3,100.00)
    (4,100.00)
    (5,100.00)
    (6,100.00)
    (7,100.00)
    (8,100.00)
    (9,100.00)
    (10,100.00)
    (11,100.00)
    (12,100.00)
    (13,100.00)
    (14,100.00)
};
\addlegendentry{GPT-6}

\addplot[
    thick,
    mark=square*,
    mark size=1.5pt,
    color=orange,
] coordinates {
    (0,42.89)
    (1,100.00)
    (2,100.00)
    (3,100.00)
    (4,100.00)
    (5,100.00)
    (6,100.00)
    (7,100.00)
    (8,100.00)
    (9,100.00)
    (10,100.00)
    (11,100.00)
    (12,100.00)
    (13,100.00)
    (14,100.00)
};
\addlegendentry{Claude Opus 5.5}

\addplot[
    thick,
    mark=triangle*,
    mark size=1.7pt,
    color=teal,
] coordinates {
    (0,50.47)
    (1,77.85)
    (2,53.09)
    (3,53.09)
    (4,53.09)
    (5,53.09)
    (6,53.09)
    (7,53.09)
    (8,53.09)
    (9,53.09)
    (10,53.56)
    (11,53.09)
    (12,66.04)
    (13,82.75)
    (14,100.00)
};
\addlegendentry{GPT-5.5 (Medium)}

\end{axis}
\end{tikzpicture}

\scriptsize
\textbf{(b) Self-Debugging}
\end{minipage}

\caption{
Trace-level accuracy across optimization iterations for \ours and
self-debugging across different models. For self-debugging, the final
observed values for GPT-6 and Claude Opus 5.5 are extended through the
remaining iterations.
}
\label{fig:iteration-accuracy-train}

\end{figure*}

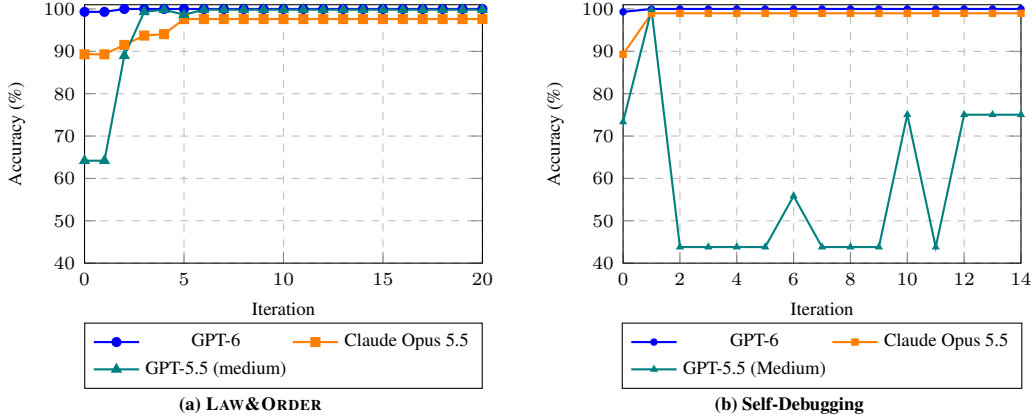
\begin{figure*}[t]
\centering

\begin{minipage}[t]{0.49\textwidth}
\centering
\begin{tikzpicture}
\begin{axis}[
    width=\linewidth,
    height=5cm,
    xlabel={Iteration},
    ylabel={Accuracy (\%)},
    xmin=0, xmax=20,
    ymin=40, ymax=101,
    xtick={0,5,10,15,20},
    ytick={40,50,60,70,80,90,100},
    grid=both,
    major grid style={dashed},
    minor grid style={dotted},
    tick label style={font=\scriptsize},
    label style={font=\scriptsize},
    legend style={
        at={(0.5,-0.23)},
        anchor=north,
        legend columns=2,
        font=\scriptsize
    },
]

\addplot[
    thick,
    mark=*,
    mark size=1.5pt,
    color=blue, 
] coordinates {
    (0,99.32) (1,99.32) (2,100.00) (3,100.00)
    (4,100.00) (5,100.00) (6,100.00) (7,100.00)
    (8,100.00) (9,100.00) (10,100.00) (11,100.00)
    (12,100.00) (13,100.00) (14,100.00) (15,100.00)
    (16,100.00) (17,100.00) (18,100.00) (19,100.00)
    (20,100.00)
};
\addlegendentry{GPT-6}

\addplot[
    thick,
    mark=square*,
    mark size=1.5pt,
    color=orange,
] coordinates {
    (0,89.30) (1,89.30) (2,91.51) (3,93.72)
    (4,94.06) (5,97.62) (6,97.62) (7,97.62)
    (8,97.62) (9,97.62) (10,97.62) (11,97.62)
    (12,97.62) (13,97.62) (14,97.62) (15,97.62)
    (16,97.62) (17,97.62) (18,97.62) (19,97.62)
    (20,97.62)
};
\addlegendentry{Claude Opus 5.5}

\addplot[
    thick,
    mark=triangle*,
    mark size=1.7pt,
    color=teal, 
] coordinates {
    (0,64.18) (1,64.18) (2,88.96) (3,99.32)
    (4,99.83) (5,98.64) (6,99.83) (7,99.83)
    (8,99.83) (9,99.83) (10,99.83) (11,99.83)
    (12,99.83) (13,99.83) (14,99.83) (15,99.83)
    (16,99.83) (17,99.83) (18,99.83) (19,99.83)
    (20,99.83)
};
\addlegendentry{GPT-5.5 (medium)}

\end{axis}
\end{tikzpicture}

\scriptsize
\textbf{(a) \ours}
\end{minipage}
\hfill
\begin{minipage}[t]{0.49\textwidth}
\centering
\begin{tikzpicture}
\begin{axis}[
    width=\linewidth,
    height=5cm,
    xlabel={Iteration},
    ylabel={Accuracy (\%)},
    xmin=0, xmax=14,
    ymin=40, ymax=101,
    xtick={0,2,4,6,8,10,12,14},
    ytick={40,50,60,70,80,90,100},
    grid=both,
    major grid style={dashed},
    minor grid style={dotted},
    tick label style={font=\scriptsize},
    label style={font=\scriptsize},
    legend style={
        at={(0.5,-0.23)},
        anchor=north,
        legend columns=2,
        font=\scriptsize
    },
]

\addplot[
    thick,
    mark=*,
    mark size=1pt,
    color=blue,
] coordinates {
    (0,99.32)
    (1,100.00)
    (2,100.00)
    (3,100.00)
    (4,100.00)
    (5,100.00)
    (6,100.00)
    (7,100.00)
    (8,100.00)
    (9,100.00)
    (10,100.00)
    (11,100.00)
    (12,100.00)
    (13,100.00)
    (14,100.00)
};
\addlegendentry{GPT-6}

\addplot[
    thick,
    mark=square*,
    mark size=1pt,
    color=orange,
] coordinates {
    (0,89.30)
    (1,98.98)
    (2,98.98)
    (3,98.98)
    (4,98.98)
    (5,98.98)
    (6,98.98)
    (7,98.98)
    (8,98.98)
    (9,98.98)
    (10,98.98)
    (11,98.98)
    (12,98.98)
    (13,98.98)
    (14,98.98)
};
\addlegendentry{Claude Opus 5.5}

\addplot[
    thick,
    mark=triangle*,
    mark size=1pt,
    color=teal,
] coordinates {
    (0,73.34)
    (1,99.83)
    (2,43.80)
    (3,43.80)
    (4,43.80)
    (5,43.80)
    (6,55.86)
    (7,43.80)
    (8,43.80)
    (9,43.80)
    (10,75.04)
    (11,43.80)
    (12,75.04)
    (13,75.04)
    (14,75.04)
};
\addlegendentry{GPT-5.5 (Medium)}

\end{axis}
\end{tikzpicture}

\scriptsize
\textbf{(b) Self-Debugging}
\end{minipage}

\caption{
Cell-level accuracy across optimization iterations for \ours and
self-debugging across different models. For self-debugging, the final
observed values for GPT-6 and Claude Opus 5.5 are extended through the
remaining iterations.
}
\label{fig:iteration-accuracy-eval}
\vspace{-8pt}
\end{figure*}

As shown in Figure~\ref{fig:iteration-accuracy-train} and Figure~\ref{fig:iteration-accuracy-eval}, in both repair and test sets, \ours shows a more stable iterative improvement than self-debugging. On the repair set, trace-level accuracy generally increases steadily under \ours, and these gains are accompanied by rapid improvements in line-level accuracy on the independently authored test set, with GPT-6 reaching 100\% and GPT-5.5 reaching 99.8\%. In contrast, although self-debugging performs well with the stronger models, it is highly unstable with GPT-5.5, exhibiting substantial regressions and fluctuations on both sets. These results indicate that \ours provides a more consistent optimization trajectory and that improvements on the repair set generally transfer to the held-out test set.
\vspace{-8pt}

\subsection{Line coverage of the test cases}
Tax forms exhibit relatively limited branching structure, allowing most cells to be exercised with only one or two examples. For Tax Year (TY) 2024, the repair set, containing at most two inputs per form, achieves 94.6\% cell coverage and 44.0\% branch coverage. By comparison, the 51-input test set increases coverage to 98.6\% of cells and 73.1\% of branches. A similar pattern holds for TY 2025: the repair set achieves 93.3\% cell coverage and 44.0\% branch coverage, whereas the 51-input test set achieves 93\% cell coverage and 73\% branch coverage. Thus, even a small number of examples exercises the vast majority of tax-form cells, although substantially more examples are required to cover alternative execution paths. Despite the sparsity of the repair sets, the relatively rigid structure and unambiguous wording of tax forms allow limited execution feedback to provide broad coverage of the underlying computations.

\vspace{-8pt}

\section{Conclusion}
\vspace{-8pt}
We proposed \ours, an iterative optimization framework for large-scale tax law auto-formalization with localized program repair. Extensive experiments and analyses demonstrate that \ours achieves strong performance beyond the human-written repair cases used for optimization, including on independently authored, CPA-written held-out tax returns across 30 tax forms. On these held-out returns, a frontier LLM correctly computes only 66\% of returns, whereas \ours achieves 100\% accuracy. Our analysis further shows that iterative repair consistently improves performance on the repair set, with corresponding improvements on held-out test cases, demonstrating the effectiveness of localized optimization for producing reliable formalizations. We hope that \ours provides a foundation for extending tax law auto-formalization to other jurisdictions, languages, and tax systems.

\section{Acknowledgement}

We would like to thank Alex Aiken, Bella Li (Clark Raymond \& Company (CPA) PLLC), Cleo Cornovardi, and Ahmad Jabbar for their fruitful discussions and constructive criticism.
\newpage
\section{Statements}
\subsection*{AI use statement}
In this work, we used generative AI tools to polish writing; generate and format tables based on our experimental results; create a flowchart of our algorithm; summarize and analyze existing literature; improve the readability of the manuscript; identify relevant literature; and format references. We also used generative AI tools to create and modify scientific figures or images, generate software code, and assist in drafting portions of the manuscript.

We did not use generative AI tools to generate synthetic datasets; propose or refine hypotheses; design or provide feedback on research methodology or experiments; assist with translation; clean or reformat datasets; conduct qualitative or thematic data analysis; interpret experimental results; develop theoretical models or conceptual frameworks; formulate mathematical claims; provide critical ingredients for proving mathematical claims; or assist in writing mathematical proofs. These uses are therefore not applicable to this work.

All AI-assisted content was reviewed by the authors. In particular, LLM-generated code was validated against real-world inputs. The authors take full responsibility for the final content of this work, including all text, claims, code, figures, and other artifacts produced with the assistance of generative AI.

\subsection*{Reproducibility statement}
We will make the data and code used in this paper publicly available upon publication. We will also release all experimental output files and provide detailed experimental procedures in the appendix.

\bibliography{iclr2026_conference}
\bibliographystyle{iclr2026_conference}

\lstset{inputencoding=utf8}
\appendix
\section{Appendix}

\subsection{Details on tax filing and tax forms. }

\paragraph{Federal income tax filing. }
The federal income tax filing process can be characterized as a sequence of interdependent computations that transform reported income into a final tax settlement. Taxpayers first aggregate income from sources such as wages, interest, dividends, retirement distributions, and Social Security benefits, after which eligible adjustments are applied to derive Adjusted Gross Income (AGI). AGI constitutes a central intermediate quantity, as numerous subsequent eligibility criteria, phase-outs, and tax computations depend on it.

Taxable income is subsequently derived from AGI by applying the relevant deductions, including the standard or itemized deduction and, where applicable, the Qualified Business Income (QBI) deduction. The resulting taxable income forms the basis for computing tax liability under the applicable IRS tax tables and worksheets, with separate computational rules governing income that receives preferential tax treatment, such as qualified dividends and capital gains. Finally, applicable tax credits and prior payments, including withholding and estimated tax payments, are applied against the resulting liability. This reconciliation determines the taxpayer's final position as either a refund or a balance due.

\begin{figure}[t]
\centering
\includegraphics[width=0.5\textwidth]{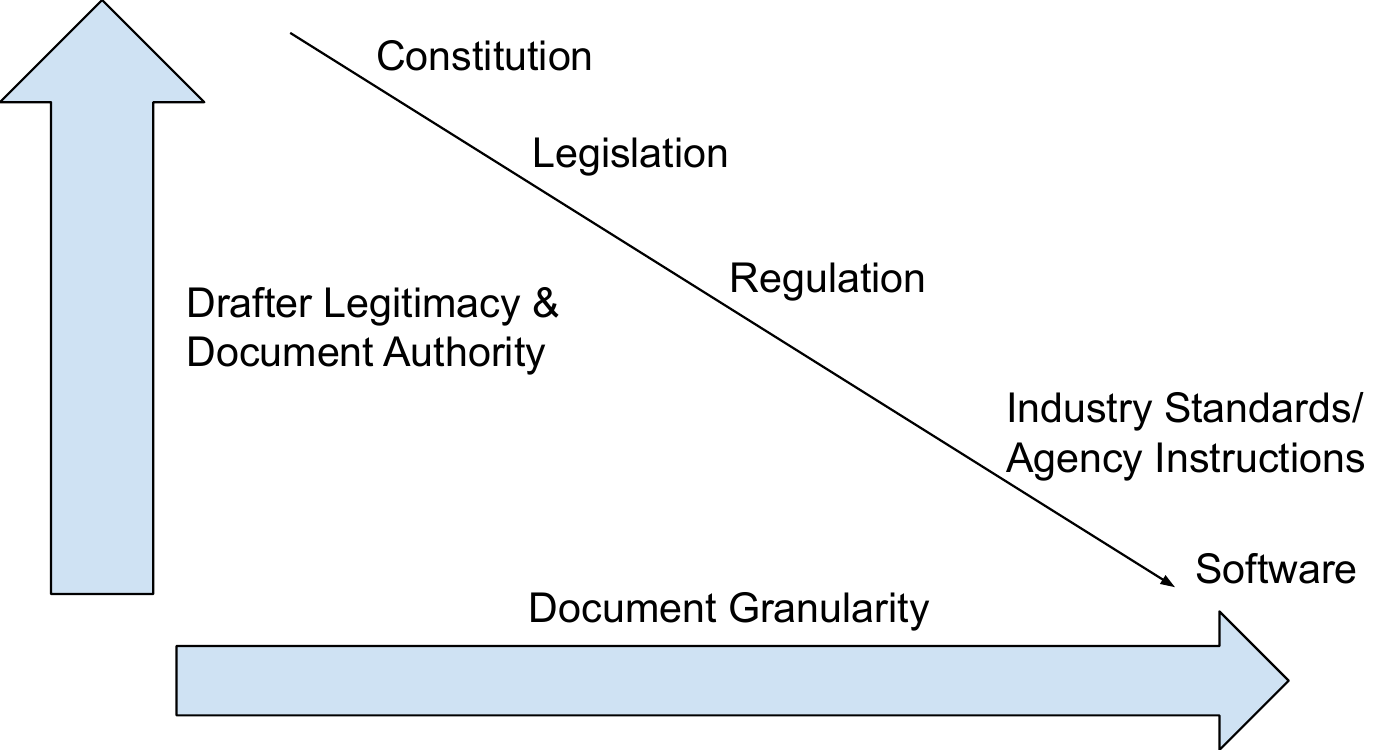}
\caption{Legal rules are built in nested refinement. A constitution has the highest authority but least detail, containing many ambiguities, vagueness, and gaps. Concrete details emerge as law is refined into more operational rules like legislation, regulation, and guidance. We formalize tax forms because they are closest to software, permitting mechanical translation.}
\label{fig:pdf}
\end{figure}

\paragraph{Progressively refined legal rules.}
As illustrated in Figure~\ref{fig:pdf}, statutes and other high-level legal sources are
progressively refined through regulations, guidance, forms, and instructions
before ultimately being implemented in software.

\paragraph{Semantic equivalence and divergence proof examples}
\label{sec:semantic_equivalence}.

\begin{figure}[t]
  \centering
  \begin{minipage}{\linewidth}
\begin{lstlisting}[style=leanfig]
-- gold
def cell_155 (s : FormInput) : Float :=
  (List.range 24).foldl
    (fun acc i => acc + schedC_add i s) 0.0
-- Claude Opus 5.5
def opus_cell_155 (s : FormInput) : Float :=
  (List.range 24).foldl
    (fun acc k => acc + schedC_add k s) 0.0

theorem eq_155 : opus_cell_155 = cell_155 := rfl
\end{lstlisting}
    \par\smallskip\centering\small (a) Equivalence
  \end{minipage}

  \bigskip

  \begin{minipage}{\linewidth}
\begin{lstlisting}[style=leanfig]
-- gold
def cell_32 (p : ParsedInput) : Nat :=
  (if over65ForVA (entryDate "YourDOB:" p) then 1 else 0) +
  (if cell_01_filing_status p == .marriedJoint &&
      over65ForVA (entryDate "SpouseDOB:" p) then 1 else 0) +
  (if cell_24_you_blind_input p then 1 else 0) +
  (if cell_01_filing_status p = .marriedJoint &&
      cell_25_spouse_blind_input p then 1 else 0)
-- Claude Opus 5.5
def opus_cell_32 (p : ParsedInput) : Nat :=
  let you := entryDate "YourDOB" p
  let sp := entryDate "SpouseDOB" p
  (if you.year > 0 && over65ForVA you then 1 else 0) +
  (if cell_24_you_blind_input p then 1 else 0) +
  (if sp.year > 0 && over65ForVA sp then 1 else 0) +
  (if cell_25_spouse_blind_input p then 1 else 0)

def input : ParsedInput :=
  parseInputText
    "Status Married/joint\nYourDOB: 6/3/1970\nSpouseDOB: 11/2/1956"

theorem neq_32 : opus_cell_32 input ≠ cell_32 input := by
  native_decide
\end{lstlisting}
    \par\smallskip\centering\small (b) Divergence
  \end{minipage}
  \caption{Lean~4 proofs relating Claude Opus 5.5's grind cells to gold.
    (a)~CA~540 Schedule CA C11--C23 total: the bodies differ only in
    bound-variable names, so \texttt{rfl} closes the goal.
    (b)~VA~760 age/blind exemptions: Opus looks up \texttt{"SpouseDOB"},
    but the input label is \texttt{"SpouseDOB:"}, so a spouse born in 1956
    is never counted as 65 or older. Gold counts 1 exemption; Opus
    counts 0, and \texttt{native\_decide} proves the outputs differ.}
  \label{fig:proofs}
\end{figure}

Most proofs of equivalence were short, using reflection tactics like Fig.~\ref{fig:proofs}(a). 

\subsection{Other analyses}
\label{sec:other_analyses}

\begin{table}[t]
\centering
\caption{Number of source files and synthesized cells for each tax form.
Across 26 forms, the dataset contains 28 source files and 1,019 synthesized cells.}
\label{tab:gold_grind_inputs}

\small
\setlength{\tabcolsep}{4pt}        
\renewcommand{\arraystretch}{0.90} 

\begin{tabular}{@{}lrr@{\hspace{0.7cm}}lrr@{}}
\toprule
\textbf{Form} & \textbf{Files} & \textbf{Cells} &
\textbf{Form} & \textbf{Files} & \textbf{Cells} \\
\midrule
CA\_540              & 1 & 153 & VA\_760             & 1 & 21 \\
Form\_CA\_5805       & 1 & 143 & MI\_1040            & 1 & 19 \\
NY\_IT201            & 1 & 118 & NC\_D400            & 1 & 19 \\
OR\_40               & 1 & 102 & Form\_2441          & 1 & 18 \\
US\_1040\_2025ty     & 2 &  76 & Form\_8959          & 1 & 18 \\
MA\_1                & 1 &  46 & Form\_8960          & 1 & 17 \\
US\_1040\_Sched\_1-A & 1 &  37 & Form\_8606          & 1 & 16 \\
AZ\_140              & 2 &  31 & Form\_8995          & 2 & 13 \\
OH\_IT1040           & 1 &  30 & Form\_2210          & 1 & 12 \\
Form\_8829           & 4 &  27 & HSA\_Form\_8889     & 1 & 12 \\
NJ\_1040             & 1 &  26 & PA\_40              & 1 & 12 \\
Form\_8812           & 1 &  21 & US\_1040\_Sched\_SE & 1 & 12 \\
Form\_6781           & 1 &   9 & US\_1040\_Sched\_C  & 1 & 11 \\
\bottomrule
\end{tabular}
\end{table}

\paragraph{Distribution of the number of cells.}
Table~\ref{tab:gold_grind_inputs} summarizes the distribution of synthesized cells across the 26 tax forms in our generated Lean code. In total, the dataset contains 28 files and 1,019 synthesized cells, with substantial variation in the number of cells across forms. The largest forms are CA 540, Form CA 5805, NY IT-201, and OR-40, containing 153, 143, 118, and 102 synthesized cells, respectively. In contrast, several smaller forms contain fewer than 20 synthesized cells. This variation reflects the heterogeneity of tax forms in the dataset, which range from relatively small schedules and auxiliary forms to larger forms requiring substantially more cell-level formalization.

\paragraph{Distribution of the number of operations for one cell.}

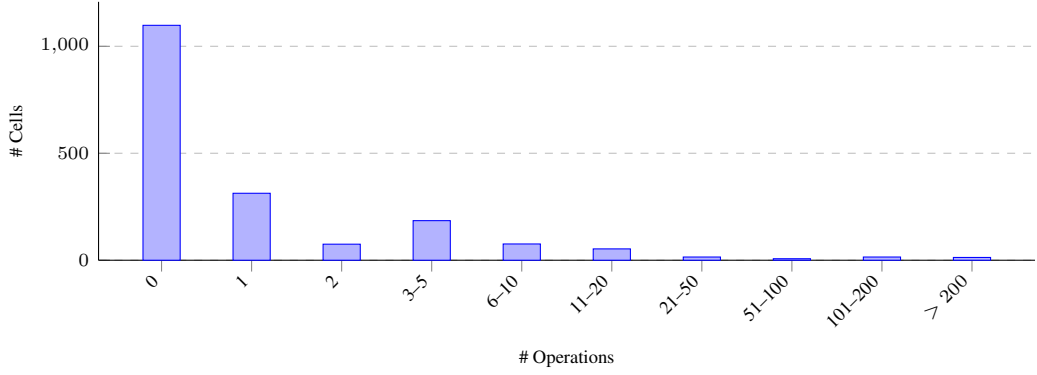
\begin{figure}[htbp]
\centering
\begin{tikzpicture}
\begin{axis}[
    ybar,
    width=\linewidth,
    height=5cm,
    bar width=14pt,
    ymin=0,
    xmin=-0.7,
    xmax=9.7,
    xlabel={\# Operations},
    ylabel={\# Cells},
    label style={font=\scriptsize},
    xtick={0,1,2,3,4,5,6,7,8,9},
    xticklabels={
        0,
        1,
        2,
        3--5,
        6--10,
        11--20,
        21--50,
        51--100,
        101--200,
        $>200$
    },
    tick label style={font=\scriptsize},
    x tick label style={
        rotate=45,
        anchor=east
    },
    ymajorgrids=true,
    grid style={dashed},
    axis x line*=bottom,
    axis y line*=left,
]
\addplot coordinates {
    (0,1098)
    (1,313)
    (2,75)
    (3,185)
    (4,76)
    (5,53)
    (6,15)
    (7,7)
    (8,15)
    (9,13)
};
\end{axis}
\end{tikzpicture}
\caption{Distribution of operator counts per
\texttt{cell\_function} across 1,847 cells from 28 tax forms. Some cells like the Taxpayer Identification Number are purely informational and thus ignored for the benchmark. Ignored cells are always zero operation.}
\label{fig:operator-distribution}
\end{figure}

Figure~\ref{fig:operator-distribution} shows the distribution of operator counts across 1,847 cells from 26 tax forms, revealing substantial variation in the computational complexity of tax-form logic. The distribution exhibits a pronounced long tail: while operator counts are generally concentrated at lower values, a nontrivial subset of cells requires substantially larger computations. Some cells contain tens or hundreds of operators, with 17 cells exceeding 200 operators. This wide range demonstrates that tax-form formalization must accommodate computations spanning several orders of magnitude in size, including large, structurally complex calculations that pose substantially greater challenges for synthesis and verification.

\paragraph{Individual forms contributing to cells with different reasoning depths.}
We show the list of individual forms contributing to cells with different reasoning depths in Table~\ref{tab:reasoning-depth-1} and Table~\ref{tab:reasoning-depth-2} below. 

\begin{table*}[t]
\centering
\small
\setlength{\tabcolsep}{6pt}
\renewcommand{\arraystretch}{1.35}
\begin{tabularx}{\textwidth}{
>{\raggedright\arraybackslash}p{0.18\textwidth}
>{\raggedright\arraybackslash}p{0.12\textwidth}
>{\raggedright\arraybackslash}X}
\rowcolor{gray!15}
\textbf{Reasoning Depth} & \textbf{\# Cells} & \textbf{Forms Contributing} \\
\hline
0 & 1021 & AZ\_140 (58), CA\_540 (92), Form\_2210 (54), Form\_2441 (35), Form\_6781 (11), Form\_8606 (20), Form\_8812 (15), Form\_8829 (34), Form\_8959 (10), Form\_8960 (25), Form\_8995 (10), Form\_CA\_5805 (79), HSA\_Form\_8889 (15), MA\_1 (55), MI\_1040 (31), NC\_D400 (21), NJ\_1040 (32), NY\_IT201 (74), OH\_IT1040 (17), OR\_40 (56), PA\_40 (24), US\_1040\_2025ty (157), US\_1040\_Sched\_1-A (17), US\_1040\_Sched\_C (46), US\_1040\_Sched\_SE (8), VA\_760 (25) \\
\hline
1 & 227 & AZ\_140 (10), CA\_540 (21), Form\_2210 (1), Form\_2441 (8), Form\_6781 (3), Form\_8606 (3), Form\_8812 (7), Form\_8829 (7), Form\_8959 (5), Form\_8960 (5), Form\_8995 (4), Form\_CA\_5805 (13), HSA\_Form\_8889 (4), MA\_1 (18), MI\_1040 (7), NC\_D400 (6), NJ\_1040 (3), NY\_IT201 (7), OH\_IT1040 (4), OR\_40 (15), PA\_40 (2), US\_1040\_2025ty (52), US\_1040\_Sched\_1-A (8), US\_1040\_Sched\_C (4), US\_1040\_Sched\_SE (3), VA\_760 (7) \\
\hline
2 & 119 & AZ\_140 (3), CA\_540 (10), Form\_2210 (2), Form\_2441 (8), Form\_6781 (1), Form\_8606 (3), Form\_8812 (4), Form\_8829 (1), Form\_8959 (4), Form\_8960 (3), Form\_8995 (4), Form\_CA\_5805 (11), HSA\_Form\_8889 (3), MA\_1 (5), MI\_1040 (3), NC\_D400 (3), NJ\_1040 (2), NY\_IT201 (2), OH\_IT1040 (2), OR\_40 (10), PA\_40 (2), US\_1040\_2025ty (23), US\_1040\_Sched\_1-A (4), US\_1040\_Sched\_C (2), US\_1040\_Sched\_SE (2), VA\_760 (2) \\
\hline
3 & 86 & AZ\_140 (2), CA\_540 (10), Form\_2210 (1), Form\_2441 (4), Form\_6781 (1), Form\_8606 (2), Form\_8812 (1), Form\_8829 (3), Form\_8959 (4), Form\_8960 (1), Form\_8995 (1), Form\_CA\_5805 (9), HSA\_Form\_8889 (2), MA\_1 (4), MI\_1040 (1), NC\_D400 (2), NJ\_1040 (2), NY\_IT201 (5), OH\_IT1040 (1), OR\_40 (5), PA\_40 (1), US\_1040\_2025ty (15), US\_1040\_Sched\_1-A (5), US\_1040\_Sched\_C (1), US\_1040\_Sched\_SE (1), VA\_760 (2) \\
\hline
4 & 55 & AZ\_140 (1), CA\_540 (3), Form\_2441 (1), Form\_6781 (1), Form\_8606 (3), Form\_8812 (1), Form\_8829 (3), Form\_8959 (2), Form\_8960 (2), Form\_8995 (1), Form\_CA\_5805 (6), HSA\_Form\_8889 (1), MA\_1 (2), MI\_1040 (1), NC\_D400 (1), NJ\_1040 (1), NY\_IT201 (3), OH\_IT1040 (1), OR\_40 (2), PA\_40 (1), US\_1040\_2025ty (11), US\_1040\_Sched\_1-A (4), US\_1040\_Sched\_C (1), US\_1040\_Sched\_SE (1), VA\_760 (1) \\
\hline
5 & 48 & AZ\_140 (1), CA\_540 (2), Form\_2441 (1), Form\_6781 (2), Form\_8606 (3), Form\_8812 (1), Form\_8829 (2), Form\_8959 (1), Form\_8960 (2), Form\_8995 (1), Form\_CA\_5805 (6), HSA\_Form\_8889 (1), MA\_1 (2), MI\_1040 (1), NC\_D400 (1), NJ\_1040 (1), NY\_IT201 (1), OH\_IT1040 (2), OR\_40 (1), PA\_40 (1), US\_1040\_2025ty (7), US\_1040\_Sched\_1-A (4), US\_1040\_Sched\_C (1), US\_1040\_Sched\_SE (2), VA\_760 (1) \\
\hline
6 & 44 & AZ\_140 (1), CA\_540 (2), Form\_2441 (1), Form\_8606 (2), Form\_8812 (1), Form\_8829 (2), Form\_8959 (1), Form\_8960 (1), Form\_CA\_5805 (5), MA\_1 (3), MI\_1040 (1), NC\_D400 (1), NJ\_1040 (4), NY\_IT201 (1), OH\_IT1040 (2), OR\_40 (1), PA\_40 (2), US\_1040\_2025ty (6), US\_1040\_Sched\_1-A (4), US\_1040\_Sched\_C (1), US\_1040\_Sched\_SE (1), VA\_760 (1) \\
\hline
7 & 35 & AZ\_140 (1), CA\_540 (1), Form\_2441 (2), Form\_8812 (1), Form\_8829 (2), Form\_8960 (1), Form\_CA\_5805 (4), MA\_1 (1), MI\_1040 (1), NC\_D400 (1), NJ\_1040 (3), NY\_IT201 (1), OH\_IT1040 (2), OR\_40 (1), PA\_40 (2), US\_1040\_2025ty (3), US\_1040\_Sched\_1-A (5), US\_1040\_Sched\_C (1), US\_1040\_Sched\_SE (1), VA\_760 (1) \\
\hline
\end{tabularx}
\caption{Distribution of cells by reasoning depth and contributing tax forms (depths 0--7).}
\label{tab:reasoning-depth-1}
\end{table*}

\begin{table*}[t]
\centering
\small
\setlength{\tabcolsep}{6pt}
\renewcommand{\arraystretch}{1.35}
\begin{tabularx}{\textwidth}{
>{\raggedright\arraybackslash}p{0.18\textwidth}
>{\raggedright\arraybackslash}p{0.12\textwidth}
>{\raggedright\arraybackslash}X}
\rowcolor{gray!15}
\textbf{Reasoning Depth} & \textbf{\# Cells} & \textbf{Forms Contributing} \\
\hline
8 & 26 & AZ\_140 (1), CA\_540 (1), Form\_2441 (2), Form\_8812 (1), Form\_8829 (1), Form\_CA\_5805 (4), MA\_1 (1), MI\_1040 (2), NC\_D400 (2), NJ\_1040 (3), NY\_IT201 (1), OH\_IT1040 (2), OR\_40 (1), US\_1040\_2025ty (2), US\_1040\_Sched\_1-A (1), VA\_760 (1) \\
\hline
9 & 24 & AZ\_140 (1), CA\_540 (1), Form\_2441 (1), Form\_8812 (1), Form\_8829 (2), Form\_CA\_5805 (4), MA\_1 (1), MI\_1040 (1), NC\_D400 (1), NJ\_1040 (2), NY\_IT201 (1), OH\_IT1040 (2), OR\_40 (1), US\_1040\_2025ty (2), US\_1040\_Sched\_1-A (1), VA\_760 (2) \\
\hline
10 & 19 & AZ\_140 (2), CA\_540 (1), Form\_2441 (1), Form\_8812 (1), Form\_8829 (1), Form\_CA\_5805 (4), MA\_1 (1), NJ\_1040 (1), NY\_IT201 (1), OH\_IT1040 (1), OR\_40 (1), US\_1040\_2025ty (2), VA\_760 (2) \\
\hline
11 & 17 & AZ\_140 (1), CA\_540 (1), Form\_2441 (1), Form\_8812 (1), Form\_CA\_5805 (4), MA\_1 (1), NJ\_1040 (2), NY\_IT201 (1), OH\_IT1040 (2), OR\_40 (1), US\_1040\_2025ty (2) \\
\hline
12 & 14 & AZ\_140 (2), CA\_540 (1), Form\_2441 (1), Form\_CA\_5805 (4), MA\_1 (1), NY\_IT201 (1), OH\_IT1040 (1), OR\_40 (1), US\_1040\_2025ty (2) \\
\hline
13 & 10 & CA\_540 (1), Form\_2441 (1), Form\_CA\_5805 (1), MA\_1 (2), NY\_IT201 (1), OR\_40 (2), US\_1040\_2025ty (2) \\
\hline
14 & 11 & CA\_540 (1), Form\_2441 (1), Form\_CA\_5805 (2), MA\_1 (1), NY\_IT201 (1), OR\_40 (2), US\_1040\_2025ty (3) \\
\hline
15 & 10 & CA\_540 (1), Form\_CA\_5805 (3), MA\_1 (1), NY\_IT201 (1), OR\_40 (1), US\_1040\_2025ty (3) \\
\hline
16 & 11 & CA\_540 (1), Form\_CA\_5805 (4), NY\_IT201 (2), OR\_40 (1), US\_1040\_2025ty (3) \\
\hline
17 & 14 & CA\_540 (1), Form\_CA\_5805 (6), NY\_IT201 (3), US\_1040\_2025ty (4) \\
\hline
18 & 10 & CA\_540 (2), Form\_CA\_5805 (4), NY\_IT201 (3), US\_1040\_2025ty (1) \\
\hline
19 & 6 & CA\_540 (1), Form\_CA\_5805 (3), NY\_IT201 (1), US\_1040\_2025ty (1) \\
\hline
20 & 7 & CA\_540 (1), Form\_CA\_5805 (5), NY\_IT201 (1) \\
\hline
21 & 8 & Form\_CA\_5805 (7), NY\_IT201 (1) \\
\hline
22 & 4 & Form\_CA\_5805 (3), NY\_IT201 (1) \\
\hline
23 & 4 & Form\_CA\_5805 (3), NY\_IT201 (1) \\
\hline
24 & 6 & Form\_CA\_5805 (4), NY\_IT201 (2) \\
\hline
25 & 5 & Form\_CA\_5805 (4), NY\_IT201 (1) \\
\hline
26 & 1 & Form\_CA\_5805 (1) \\
\hline
27 & 1 & Form\_CA\_5805 (1) \\
\hline
28 & 2 & Form\_CA\_5805 (2) \\
\hline
29 & 1 & Form\_CA\_5805 (1) \\
\hline
30 & 1 & Form\_CA\_5805 (1) \\
\hline
\end{tabularx}
\caption{Distribution of cells by reasoning depth and contributing tax forms (depths 8--30).}
\label{tab:reasoning-depth-2}
\end{table*}

\subsection{Experimental details}
\paragraph{Hyper-parameters.} We set a maximum of 20 iterations for both the self-debugging and \ours runs.

\paragraph{Prompts.} We show the prompts used in different stages of our algorithm below. 


\begin{longtable}{@{}>{\raggedright\arraybackslash\ttfamily\footnotesize}p{\linewidth}@{}}
\caption{Translation prompt (\texttt{TRANSLATION\_PROMPT}). The placeholders \textit{method\_text}, \textit{hint\_text}, \textit{cell\_breakdown}, \textit{section\_parser\_snippet}, and \textit{row\_pattern\_snippet} are expanded as shown in Table~\ref{tab:prompt-method-instructions}--Table~\ref{tab:prompt-row-pattern-snippet}.}\label{tab:prompt-translation}\\
\toprule
\endfirsthead
\multicolumn{1}{@{}l@{}}{\small\itshape Table~\thetable{} (continued)}\\
\toprule
\endhead
\midrule
\multicolumn{1}{@{}r@{}}{\small\itshape Continued on next page}\\
\endfoot
\bottomrule
\endlastfoot
You complete a Lean 4 tax-form skeleton (\textasciigrave{}Template.lean\textasciigrave{}). \\
 \\
The skeleton already defines helpers, structures, \textasciigrave{}render\textasciigrave{}, and \textasciigrave{}main\textasciigrave{}. \\
Each \textasciigrave{}def cell\_*\textasciigrave{} function whose body is \textasciigrave{}sorry\textasciigrave{} must be filled in using the \\
form instructions provided below. These include the instruction booklet (when \\
bundled), inline \textasciigrave{}\{...\}\textasciigrave{} guidance from the form template, and \textasciigrave{}--\textasciigrave{} comments on \\
cells in \textasciigrave{}Template.lean\textasciigrave{}. \\
Do not change function names, type signatures, or the overall file structure except \\
to replace \textasciigrave{}sorry\textasciigrave{} bodies with real implementations. \\
Leave any \textasciigrave{}def cell\_*\textasciigrave{} that already reads an input (for example a single \textasciigrave{}inputNum\textasciigrave{} \\
or \textasciigrave{}entryNumber\textasciigrave{} call) unchanged. \\
 \\
Return only one complete Lean 4 file inside a single \textasciigrave{}\textasciigrave{}\textasciigrave{}lean\textasciigrave{}\textasciigrave{}\textasciigrave{} code block. \\
Do not write prose before or after the code block. \\
The file will be checked with \textasciigrave{}lake env lean --run\textasciigrave{} under Lean 4.29.1. \\
Do not use C code, gold outputs, JSON, or APIs that are not imported or defined. \\
The first non-comment command in the file must be exactly \textasciigrave{}import ExampleParse\textasciigrave{}. \\
Do not put \textasciigrave{}set\_option\textasciigrave{}, \textasciigrave{}open\textasciigrave{}, \textasciigrave{}namespace\textasciigrave{}, or any declaration before that import. \\
Do not import \textasciigrave{}Mathlib\textasciigrave{}, \textasciigrave{}Std\textasciigrave{}, or any module other than \textasciigrave{}ExampleParse\textasciigrave{}. \\
The submitted file must not contain \textasciigrave{}sorry\textasciigrave{}. \\
 \\
Portable ExampleParse contract: \\
- Every benchmark provides \textasciigrave{}parseInputFile (path : System.FilePath) : IO ParsedInput\textasciigrave{}. \\
- Every benchmark provides \textasciigrave{}ParsedInput.entries : List Entry\textasciigrave{}, and each \textasciigrave{}Entry\textasciigrave{} has \textasciigrave{}label : String\textasciigrave{} \\
\hspace*{1.050em}and \textasciigrave{}values : List String\textasciigrave{}. \\
- The common helper subset across benchmarks is: \\
\hspace*{1.050em}\textasciigrave{}entryValues\textasciigrave{}, \textasciigrave{}entryText\textasciigrave{}, \textasciigrave{}entryNumber\textasciigrave{}, \textasciigrave{}cleanLine\textasciigrave{}, \textasciigrave{}words\textasciigrave{}, \textasciigrave{}trimString\textasciigrave{}, \textasciigrave{}toFloat?\textasciigrave{}. \\
- On the shared parser used here, helper return types are: \\
\hspace*{1.050em}\textasciigrave{}entryValues : String -> ParsedInput -> List String\textasciigrave{}, \\
\hspace*{1.050em}\textasciigrave{}entryText : String -> ParsedInput -> String\textasciigrave{}, \\
\hspace*{1.050em}\textasciigrave{}entryNumber : String -> ParsedInput -> Float\textasciigrave{}, \\
\hspace*{1.050em}\textasciigrave{}entryInt : String -> ParsedInput -> Nat\textasciigrave{}, \\
\hspace*{1.050em}\textasciigrave{}entryBool : String -> ParsedInput -> Bool\textasciigrave{}. \\
- Some benchmarks also define extra helpers such as \textasciigrave{}entryBool\textasciigrave{}, \textasciigrave{}entryInt\textasciigrave{}, \textasciigrave{}statusFromInput\textasciigrave{}, \\
\hspace*{1.050em}\textasciigrave{}entityFromInput\textasciigrave{}, \textasciigrave{}entryNumberAt\textasciigrave{}, or \textasciigrave{}ParsedInput.raw\textasciigrave{}, but many do not. \\
- Use prefix calls such as \textasciigrave{}entryText "YourName:" p\textasciigrave{}, \textasciigrave{}toFloat? tok\textasciigrave{}, \textasciigrave{}trimString s\textasciigrave{}. \\
- The \textasciigrave{}entry*\textasciigrave{} helpers take the whole \textasciigrave{}ParsedInput\textasciigrave{}, not \textasciigrave{}p.entries\textasciigrave{}. \\
\hspace*{1.050em}If you bind \textasciigrave{}let entries := p.entries\textasciigrave{}, do not pass \textasciigrave{}entries\textasciigrave{} into \textasciigrave{}entryText\textasciigrave{} or \textasciigrave{}entryNumber\textasciigrave{}. \\
- Do not pattern-match on \textasciigrave{}entryText\textasciigrave{}, \textasciigrave{}entryNumber\textasciigrave{}, \textasciigrave{}entryInt\textasciigrave{}, or \textasciigrave{}entryBool\textasciigrave{} as if they returned \textasciigrave{}Option\textasciigrave{}. \\
\hspace*{1.050em}If you need an optional parse, use \textasciigrave{}toFloat?\textasciigrave{} on raw strings from \textasciigrave{}entryValues\textasciigrave{}. \\
- The only numeric parser you should assume exists is \textasciigrave{}toFloat? (s : String) : Option Float\textasciigrave{}. \\
\hspace*{1.050em}There is no \textasciigrave{}String.toFloat?\textasciigrave{}, \textasciigrave{}Float.toLong\textasciigrave{}, or \textasciigrave{}Float.toInt\textasciigrave{}. \\
- If helper availability is uncertain, write a small local helper over \textasciigrave{}p.entries\textasciigrave{} or \textasciigrave{}entryText\textasciigrave{} \\
\hspace*{1.050em}instead of assuming a benchmark-specific API exists. \\
- For checkbox logic, prefer a fresh local helper such as \textasciigrave{}localTruthy : String -> Bool\textasciigrave{} \\
\hspace*{1.050em}over assuming \textasciigrave{}entryBool\textasciigrave{} exists. \\
- Do not redefine names that may already be imported from \textasciigrave{}ExampleParse\textasciigrave{}, such as \textasciigrave{}truthy\textasciigrave{}, \\
\hspace*{1.050em}\textasciigrave{}cleanLine\textasciigrave{}, \textasciigrave{}words\textasciigrave{}, \textasciigrave{}trimString\textasciigrave{}, \textasciigrave{}toFloat?\textasciigrave{}, \textasciigrave{}entryText\textasciigrave{}, \textasciigrave{}entryValues\textasciigrave{}, \textasciigrave{}entryNumber\textasciigrave{}, \\
\hspace*{1.050em}\textasciigrave{}entryBool\textasciigrave{}, \textasciigrave{}entryInt\textasciigrave{}, \textasciigrave{}statusFromInput\textasciigrave{}, or \textasciigrave{}entityFromInput\textasciigrave{}. \\
\hspace*{1.050em}If you need a local variant, give it a fresh name like \textasciigrave{}localTruthy\textasciigrave{}. \\
\hspace*{1.050em}Avoid new top-level names that collide with common field names such as \textasciigrave{}entries\textasciigrave{}, \textasciigrave{}values\textasciigrave{}, or \textasciigrave{}label\textasciigrave{}. \\
- If \textasciigrave{}statusFromInput\textasciigrave{} or \textasciigrave{}entityFromInput\textasciigrave{} are already available from \textasciigrave{}ExampleParse\textasciigrave{}, use them directly \\
\hspace*{1.050em}instead of redefining them. \\
- Do not write \textasciigrave{}ExampleParse.parseInputFile\textasciigrave{}, \textasciigrave{}tok.toFloat?\textasciigrave{}, \textasciigrave{}s.trim\textasciigrave{}, parser combinators, \\
\hspace*{1.050em}guessed helpers like \textasciigrave{}getFloat!\textasciigrave{}, or guessed utilities like \textasciigrave{}List.enum\textasciigrave{}. \\
- Do not use \textasciigrave{}List.enum\textasciigrave{}, \textasciigrave{}List.enumFrom\textasciigrave{}, \textasciigrave{}mapIdx\textasciigrave{}, \textasciigrave{}List.mapIdx\textasciigrave{}, \textasciigrave{}List.join\textasciigrave{}, \textasciigrave{}mut\textasciigrave{}, \textasciigrave{}String.trim\textasciigrave{}, \\
\hspace*{1.050em}bare \textasciigrave{}getD\textasciigrave{}, or \textasciigrave{}Float.toInt\textasciigrave{}. \\
\hspace*{1.050em}If you need row indices, write a recursive helper that threads \textasciigrave{}idx : Nat\textasciigrave{}. \\
\hspace*{1.050em}If you need a default from \textasciigrave{}Option\textasciigrave{}, use pattern matching or \textasciigrave{}opt.getD default\textasciigrave{}. \\
\hspace*{1.050em}Write the full application, for example \textasciigrave{}(toFloat? s).getD 0.0\textasciigrave{}. \\
\hspace*{1.050em}Do not use proof-based indexing such as \textasciigrave{}xs[i]\textasciigrave{}, \textasciigrave{}List.nthLe\textasciigrave{}, or \textasciigrave{}xs.get $\langle$i, h$\rangle$\textasciigrave{}. \\
\hspace*{1.050em}After \textasciigrave{}splitOn\textasciigrave{}, pattern match on the resulting list instead of using \textasciigrave{}get!\textasciigrave{}. \\
\hspace*{1.050em}Example: \textasciigrave{}match s.splitOn "-" with | a :: b :: c :: \_ => ... | \_ => ...\textasciigrave{}. \\
\hspace*{1.050em}Use \textasciigrave{}String.intercalate\textasciigrave{} and \textasciigrave{}trimString s\textasciigrave{}, not dot notation like \textasciigrave{}xs.intercalate\textasciigrave{} or \textasciigrave{}s.trimString\textasciigrave{}. \\
 \\
Program contract: \\
- Import \textasciigrave{}ExampleParse\textasciigrave{}. \\
- Fill every \textasciigrave{}def cell\_*\textasciigrave{} whose body is currently \textasciigrave{}sorry\textasciigrave{} according to the form \\
\hspace*{1.050em}instructions below (not by guessing from examples or gold outputs). \\
- Input cells should read from \textasciigrave{}ParsedInput\textasciigrave{} (or \textasciigrave{}FormState\textasciigrave{} when the skeleton does); \\
\hspace*{1.050em}computed cells should be formulas over earlier \textasciigrave{}cell\_*\textasciigrave{} functions. \\
- Keep the existing \textasciigrave{}render\textasciigrave{} and \textasciigrave{}main\textasciigrave{} definitions unless a minimal edit is required \\
\hspace*{1.050em}for your filled \textasciigrave{}cell\_*\textasciigrave{} functions to typecheck. \\
- Reuse only real form labels. Do not emit notes, comments, section headers, \textasciigrave{}Part\textasciigrave{}, or synthetic labels. \\
- Default strategy for most forms: use \textasciigrave{}entryText\textasciigrave{}, \textasciigrave{}entryNumber\textasciigrave{}, local helpers over \textasciigrave{}p.entries\textasciigrave{}, \\
\hspace*{1.050em}and direct scalar formulas. Most benchmarks are scalar forms and do not need custom row parsing. \\
- Some visible labels start repeating tables or worksheets rather than scalar cells. Do not emit \\
\hspace*{1.050em}\textasciigrave{}L1 = ...\textasciigrave{} just because a section starts with \textasciigrave{}L1\textasciigrave{}; treat it as rows, not one scalar cell. \\
- Only use \textasciigrave{}p.raw\textasciigrave{} or raw-line section parsing on benchmarks that actually expose \textasciigrave{}ParsedInput.raw\textasciigrave{}. \\
\hspace*{1.050em}Many forms do not expose \textasciigrave{}raw\textasciigrave{}, so on those forms you must stay within \textasciigrave{}p.entries\textasciigrave{} and the common helpers. \\
- A template likely needs raw row parsing only when a label is followed by continuation lines up to \textasciigrave{};\textasciigrave{} \\
\hspace*{1.050em}and simple \textasciigrave{}entryValues\textasciigrave{} would lose row boundaries. Otherwise avoid raw parsing. \\
- For repeating rows, use labels like \textasciigrave{}L10row1\_a\textasciigrave{}, \textasciigrave{}L10row1\_b\textasciigrave{}, ... with the form's row order and \\
\hspace*{1.050em}column-letter suffixes. Do not invent descriptive suffixes like \textasciigrave{}\_gross\textasciigrave{}, \textasciigrave{}\_date\textasciigrave{}, or \textasciigrave{}\_loss\textasciigrave{}. \\
- If a row line contains brace text like \textasciigrave{}\{ Broker A \}\textasciigrave{}, that brace text is usually the text cell. \\
- When a repeated row has a description or date text column, that column must use \textasciigrave{}textLine\textasciigrave{}, \\
\hspace*{1.050em}not numeric parsing. \\
- Illustrative example only: \\
\hspace*{1.050em}wrong: \textasciigrave{}L1 = 1710.99\textasciigrave{}, \textasciigrave{}L10row1\_gross = 1000.0\textasciigrave{} \\
\hspace*{1.050em}right: \textasciigrave{}L1row1\_a: Broker A\textasciigrave{}, \textasciigrave{}L1row1\_b = -300.99\textasciigrave{}, \textasciigrave{}L1row2\_c = 10.99\textasciigrave{}, \textasciigrave{}L10row1\_d = 1000.0\textasciigrave{} \\
- Emit all present rows in order, capped by the visible row count of the form. \\
- Use simple recursive helpers and list pattern matching such as \textasciigrave{}a :: b :: rest\textasciigrave{}. \\
\hspace*{1.050em}Avoid \textasciigrave{}List.get!\textasciigrave{}, \textasciigrave{}get?\textasciigrave{}, imperative loops, and parser DSLs. \\
- \textasciigrave{}entryValues\textasciigrave{} is fine for simple cells but loses row boundaries for multiline tables. \\
\hspace*{1.050em}Parse repeating sections from \textasciigrave{}p.raw\textasciigrave{} only on benchmarks that actually define \textasciigrave{}raw\textasciigrave{}. \\
- Parse repeating rows first, then compute downstream scalar cells from those parsed rows. \\
- If and only if the benchmark exposes \textasciigrave{}p.raw\textasciigrave{}, a good repeated-section pattern is: \\
\textit{\{section\_parser\_snippet\}} \\
- If and only if the benchmark exposes \textasciigrave{}p.raw\textasciigrave{}, a good row-building and row-rendering pattern is: \\
\textit{\{row\_pattern\_snippet\}} \\
- Emit deterministic trace lines only: \\
\hspace*{1.050em}- numeric cells as \textasciigrave{}Label = value\textasciigrave{} \\
\hspace*{1.050em}- text and checkbox cells as \textasciigrave{}Label: value\textasciigrave{} \\
\hspace*{1.050em}- checkbox values as \textasciigrave{}X\textasciigrave{} or empty text, never \textasciigrave{}true\textasciigrave{} or \textasciigrave{}false\textasciigrave{} \\
- Numeric input cells still render with \textasciigrave{}line\textasciigrave{}, not \textasciigrave{}textLine\textasciigrave{}. \\
- Omit optional row cells that should be absent in the comparable trace instead of printing \textasciigrave{}= 0\textasciigrave{}. \\
- The checker compares emitted cells in order. An early wrong label causes an immediate failure, \\
\hspace*{1.050em}so do not emit approximate summaries before the exact row/cell labels. \\
- Prefer small helpers like: \\
\hspace*{1.050em}\textasciigrave{}def fmt (x : Float) : String := toString x\textasciigrave{} \\
\hspace*{1.050em}\textasciigrave{}def line (label : String) (value : Float) : String := label ++ " = " ++ fmt value\textasciigrave{} \\
\hspace*{1.050em}\textasciigrave{}def textLine (label value : String) : String := label ++ ": " ++ value\textasciigrave{} \\
\hspace*{1.050em}\textasciigrave{}def notLessThanZero (x : Float) : Float := if x < 0.0 then 0.0 else x\textasciigrave{} \\
- Row renderers should return \textasciigrave{}List String\textasciigrave{}, not one newline-joined \textasciigrave{}String\textasciigrave{}. \\
\hspace*{1.050em}Use the recursive \textasciigrave{}renderIndexedFrom\textasciigrave{} style above instead of \textasciigrave{}mapIdx\textasciigrave{}. \\
- If you normalize dates, split the original date string and use \textasciigrave{}String.toNat?\textasciigrave{}, not float arithmetic. \\
- End with \textasciigrave{}render : ParsedInput -> String\textasciigrave{} and this exact main shape: \\
 \\
\textasciigrave{}\textasciigrave{}\textasciigrave{}lean \\
def main (args : List String) : IO UInt32 := do \\
\hspace*{1.050em}match args with \\
\hspace*{1.050em}| inputPath :: \_ => \\
\hspace*{3.150em}let parsed <- parseInputFile inputPath \\
\hspace*{3.150em}IO.print (render parsed) \\
\hspace*{3.150em}return 0 \\
\hspace*{1.050em}| [] => \\
\hspace*{3.150em}IO.eprintln "usage: form <example input>" \\
\hspace*{3.150em}return 1 \\
\textasciigrave{}\textasciigrave{}\textasciigrave{} \\
 \\
- Avoid Lean warnings and unused helpers because warning text breaks trace comparison. \\
- The checker effectively treats warnings as failures. Never use deprecated APIs such as \textasciigrave{}String.trim\textasciigrave{}. \\
- It is acceptable to add \textasciigrave{}set\_option linter.unusedVariables false\textasciigrave{} near the top if needed. \\
- Keep the code conservative and typecheck-first. \\
\textit{\{method\_text\}}\textit{\{hint\_text\}}\textit{\{cell\_breakdown\}} \\
Form instructions (booklet + inline form text; use these to fill each \textasciigrave{}sorry\textasciigrave{}): \\
\textit{\{form\_instructions\}} \\
 \\
Lean skeleton (\textasciigrave{}Template.lean\textasciigrave{}): \\
\textit{\{tax\_form\_template\}} \\
\end{longtable}

\begin{longtable}{@{}>{\raggedright\arraybackslash\ttfamily\footnotesize}p{\linewidth}@{}}
\caption{Method instructions (\texttt{METHOD\_INSTRUCTIONS}), substituted for \textit{method\_text} in the translation prompt. Empty in the ablation.}\label{tab:prompt-method-instructions}\\
\toprule
\endfirsthead
\multicolumn{1}{@{}l@{}}{\small\itshape Table~\thetable{} (continued)}\\
\toprule
\endhead
\midrule
\multicolumn{1}{@{}r@{}}{\small\itshape Continued on next page}\\
\endfoot
\bottomrule
\endlastfoot
Use a deliberate completion method: \\
1. Read the form instructions provided below (instruction booklet, inline \textasciigrave{}\{...\}\textasciigrave{} \\
\hspace*{1.575em}notes from the form template, and \textasciigrave{}--\textasciigrave{} comments in Template.lean), then the \\
\hspace*{1.575em}skeleton helpers and existing input \textasciigrave{}cell\_*\textasciigrave{} functions. \\
2. For every \textasciigrave{}def cell\_*\textasciigrave{} whose body is \textasciigrave{}sorry\textasciigrave{}, implement the computation described \\
\hspace*{1.575em}in those instructions for that line or cell. Match each function to the form using \\
\hspace*{1.575em}its name, labels, and comments in Template.lean. \\
3. Fill sorries in dependency order so later cells can call earlier ones. \\
4. Encode each computed cell as a small formula over other \textasciigrave{}cell\_*\textasciigrave{} functions and \\
\hspace*{1.575em}inputs. Do not invent formulas that contradict the instructions. \\
5. Keep rendering deterministic and ordered like the form. \\
6. Prefer simple Float arithmetic and small helper functions over hard-coded examples. \\
\end{longtable}

\begin{longtable}{@{}>{\raggedright\arraybackslash\ttfamily\footnotesize}p{\linewidth}@{}}
\caption{Repair-round hint text, substituted for \textit{hint\_text} in the translation prompt when feedback from a previous attempt is available.}\label{tab:prompt-hint-text}\\
\toprule
\endfirsthead
\multicolumn{1}{@{}l@{}}{\small\itshape Table~\thetable{} (continued)}\\
\toprule
\endhead
\midrule
\multicolumn{1}{@{}r@{}}{\small\itshape Continued on next page}\\
\endfoot
\bottomrule
\endlastfoot
Highest-priority repair target from the previous attempt: \\
\textit{\{hints\}} \\
Use that feedback as the first thing to fix. \\
Preserve earlier cells that already match. \\
If the feedback names a first failing cell, keep every earlier emitted cell unchanged and make the next emitted output line match the expected cell exactly before changing later cells. \\
If the feedback is a Lean diagnostic, fix that exact construct first and avoid unrelated rewrites. \\
\textit{\{available\_cells\_section\}} \\
\end{longtable}

\begin{longtable}{@{}>{\raggedright\arraybackslash\ttfamily\footnotesize}p{\linewidth}@{}}
\caption{Available-cells section, substituted for \textit{available\_cells\_section} in the hint text and the repair prompt.}\label{tab:prompt-available-cells}\\
\toprule
\endfirsthead
\multicolumn{1}{@{}l@{}}{\small\itshape Table~\thetable{} (continued)}\\
\toprule
\endhead
\midrule
\multicolumn{1}{@{}r@{}}{\small\itshape Continued on next page}\\
\endfoot
\bottomrule
\endlastfoot
Available cell functions (use only these identifiers; do not invent new cell\_* names or guess namespaced spellings): \\
\textit{\{joined\}} \\
\end{longtable}

\begin{longtable}{@{}>{\raggedright\arraybackslash\ttfamily\footnotesize}p{\linewidth}@{}}
\caption{Cell-breakdown header (\texttt{CELL\_BREAKDOWN\_HEADER}), prepended to the per-cell instruction text substituted for \textit{cell\_breakdown}.}\label{tab:prompt-cell-breakdown}\\
\toprule
\endfirsthead
\multicolumn{1}{@{}l@{}}{\small\itshape Table~\thetable{} (continued)}\\
\toprule
\endhead
\midrule
\multicolumn{1}{@{}r@{}}{\small\itshape Continued on next page}\\
\endfoot
\bottomrule
\endlastfoot
Per-cell breakdown of the \textasciigrave{}sorry\textasciigrave{} functions you must implement. \\
Each entry gives the \textasciigrave{}def cell\_*\textasciigrave{} you must fill and the form text describing it. \\
Implement exactly these functions; leave every other definition unchanged. \\
\end{longtable}

\begin{longtable}{@{}>{\raggedright\arraybackslash\ttfamily\footnotesize}p{\linewidth}@{}}
\caption{Section-parser snippet (\texttt{SECTION\_PARSER\_SNIPPET}), substituted for \textit{section\_parser\_snippet} in the translation prompt.}\label{tab:prompt-section-parser-snippet}\\
\toprule
\endfirsthead
\multicolumn{1}{@{}l@{}}{\small\itshape Table~\thetable{} (continued)}\\
\toprule
\endhead
\midrule
\multicolumn{1}{@{}r@{}}{\small\itshape Continued on next page}\\
\endfoot
\bottomrule
\endlastfoot
\textasciigrave{}\textasciigrave{}\textasciigrave{}lean \\
def wordsBeforeSemicolon (tokens : List String) : List String := \\
\hspace*{1.050em}tokens.takeWhile (fun tok => tok != ";") \\
 \\
def startsSection (label : String) (line : String) : Bool := \\
\hspace*{1.050em}match words (cleanLine line) with \\
\hspace*{1.050em}| first :: \_ => first = label \\
\hspace*{1.050em}| [] => false \\
 \\
partial def collectSectionLines (label : String) (lines : List String) (started : Bool) (acc : List String) : List String := \\
\hspace*{1.050em}match lines with \\
\hspace*{1.050em}| [] => acc.reverse \\
\hspace*{1.050em}| line :: rest => \\
\hspace*{3.150em}if started then \\
\hspace*{4.200em}let acc' := line :: acc \\
\hspace*{4.200em}if (words (cleanLine line)).any (fun tok => tok = ";") then acc'.reverse \\
\hspace*{4.200em}else collectSectionLines label rest true acc' \\
\hspace*{3.150em}else if startsSection label line then \\
\hspace*{4.200em}let acc' := [line] \\
\hspace*{4.200em}if (words (cleanLine line)).any (fun tok => tok = ";") then acc' \\
\hspace*{4.200em}else collectSectionLines label rest true acc' \\
\hspace*{3.150em}else collectSectionLines label rest false acc \\
 \\
def sectionLines (label : String) (p : ParsedInput) : List String := \\
\hspace*{1.050em}collectSectionLines label (p.raw.splitOn "\textbackslash{}n") false [] \\
 \\
def braceText (line : String) : Option String := \\
\hspace*{1.050em}match line.splitOn "\{" with \\
\hspace*{1.050em}| \_ :: after :: \_ => \\
\hspace*{3.150em}match after.splitOn "\}" with \\
\hspace*{3.150em}| inside :: \_ => some (trimString inside) \\
\hspace*{3.150em}| [] => none \\
\hspace*{1.050em}| \_ => none \\
\textasciigrave{}\textasciigrave{}\textasciigrave{} \\
\end{longtable}

\begin{longtable}{@{}>{\raggedright\arraybackslash\ttfamily\footnotesize}p{\linewidth}@{}}
\caption{Row-pattern snippet (\texttt{ROW\_PATTERN\_SNIPPET}), substituted for \textit{row\_pattern\_snippet} in the translation prompt.}\label{tab:prompt-row-pattern-snippet}\\
\toprule
\endfirsthead
\multicolumn{1}{@{}l@{}}{\small\itshape Table~\thetable{} (continued)}\\
\toprule
\endhead
\midrule
\multicolumn{1}{@{}r@{}}{\small\itshape Continued on next page}\\
\endfoot
\bottomrule
\endlastfoot
\textasciigrave{}\textasciigrave{}\textasciigrave{}lean \\
structure Row where \\
\hspace*{1.050em}desc : String \\
\hspace*{1.050em}x : Float \\
\hspace*{1.050em}y : Float \\
 \\
partial def buildRows : List String -> List String -> List Row \\
\hspace*{1.050em}| desc :: descs, x :: y :: rest => \\
\hspace*{3.150em}\{ desc := desc, x := (toFloat? x).getD 0.0, y := (toFloat? y).getD 0.0 \} :: \\
\hspace*{4.200em}buildRows descs rest \\
\hspace*{1.050em}| \_, \_ => [] \\
 \\
def renderRow (idx : Nat) (row : Row) : List String := \\
\hspace*{1.050em}[textLine ("L1row" ++ toString idx ++ "\_a") row.desc, \\
\hspace*{1.575em}if row.x != 0.0 then line ("L1row" ++ toString idx ++ "\_b") row.x else "", \\
\hspace*{1.575em}if row.y != 0.0 then line ("L1row" ++ toString idx ++ "\_c") row.y else ""] \\
 \\
partial def renderIndexedFrom \{$\alpha$ : Type\} (limit : Nat) (idx : Nat) \\
\hspace*{2.100em}(renderRow : Nat -> $\alpha$ -> List String) : List $\alpha$ -> List String \\
\hspace*{1.050em}| [] => [] \\
\hspace*{1.050em}| row :: rows => \\
\hspace*{3.150em}if idx > limit then [] \\
\hspace*{3.150em}else renderRow idx row ++ renderIndexedFrom limit (idx + 1) renderRow rows \\
\textasciigrave{}\textasciigrave{}\textasciigrave{} \\
\end{longtable}

\begin{longtable}{@{}>{\raggedright\arraybackslash\ttfamily\footnotesize}p{\linewidth}@{}}
\caption{Entailment prompt (\texttt{ENTAILMENT\_PROMPT}): the model fills in the computed cells of a partially filled form.}\label{tab:prompt-entailment}\\
\toprule
\endfirsthead
\multicolumn{1}{@{}l@{}}{\small\itshape Table~\thetable{} (continued)}\\
\toprule
\endhead
\midrule
\multicolumn{1}{@{}r@{}}{\small\itshape Continued on next page}\\
\endfoot
\bottomrule
\endlastfoot
Fill the missing computed cells in this tax form and return only a comparable trace. \\
 \\
Return plain text only. \\
Do not use markdown, code fences, explanations, or notes. \\
Do not emit section headers, comments, or synthetic summary labels. \\
 \\
Trace rules: \\
- Use \textasciigrave{}Label = value\textasciigrave{} for numeric cells. \\
- Use \textasciigrave{}Label: value\textasciigrave{} for text and checkbox cells. \\
- Preserve the form's order. \\
- Copy existing input cells into the trace too. \\
- Reuse only real form labels. \\
- For repeating tables, emit row/cell labels such as \textasciigrave{}L1row1\_a\textasciigrave{}, \textasciigrave{}L1row1\_b\textasciigrave{}, \textasciigrave{}L10row1\_d\textasciigrave{}, \textasciigrave{}L14row2\_e\textasciigrave{}. \\
- Use the form's row order and column-letter suffixes instead of descriptive suffixes. \\
- If a row contains brace text like \textasciigrave{}\{ Broker A \}\textasciigrave{}, use that text for the row's text cell. \\
- Wrong: \textasciigrave{}L1 = 1710.99\textasciigrave{}, \textasciigrave{}L10row1\_gross = 1000.0\textasciigrave{}. \\
\hspace*{1.050em}Right: \textasciigrave{}L1row1\_a: Broker A\textasciigrave{}, \textasciigrave{}L1row1\_b = -300.99\textasciigrave{}, \textasciigrave{}L10row1\_d = 1000.0\textasciigrave{}. \\
- Checkbox values must be \textasciigrave{}X\textasciigrave{} or empty text, never \textasciigrave{}true\textasciigrave{} or \textasciigrave{}false\textasciigrave{}. \\
- Omit optional row cells that should be absent in the comparable trace instead of printing \textasciigrave{}= 0\textasciigrave{}. \\
 \\
Partial tax form with missing computed cells: \\
\textit{\{partial\_tax\_form\}} \\
\end{longtable}

\begin{longtable}{@{}>{\raggedright\arraybackslash\ttfamily\footnotesize}p{\linewidth}@{}}
\caption{Cell-extraction prompt (\texttt{CELL\_EXTRACTION\_PROMPT}).}\label{tab:prompt-cell-extraction}\\
\toprule
\endfirsthead
\multicolumn{1}{@{}l@{}}{\small\itshape Table~\thetable{} (continued)}\\
\toprule
\endhead
\midrule
\multicolumn{1}{@{}r@{}}{\small\itshape Continued on next page}\\
\endfoot
\bottomrule
\endlastfoot
You read filled-in examples of one tax form and list every cell on the form. \\
 \\
A "cell" is one numbered line or labeled entry on the form, for example \textasciigrave{}L1\textasciigrave{}, \textasciigrave{}L4\textasciigrave{}, \textasciigrave{}BoxA\textasciigrave{}, \\
or \textasciigrave{}YourName\textasciigrave{}. For each cell decide whether it is an *input* (a value the filer enters \\
directly, copied from a document or another form) or *computed* (a value derived by \\
arithmetic from other cells on this same form). \\
 \\
Return only a single JSON array. Do not write prose, headings, or any code fence other than \\
the JSON. Each array element is an object with exactly these keys: \\
- "number": integer position of the cell on the form, starting at 1 and increasing in form order. \\
- "name": short snake\_case summary of the cell, ascii letters/digits/underscores only, no spaces, \\
\hspace*{1.050em}for example "l4\_adjustments" or "boxA\_mixed\_straddle". \\
- "label": the exact label as written in the form text, for example "L4", "BoxA", "YourName". \\
- "kind": either "input" or "computed". \\
- "value\_kind": one of "number", "text", "checkbox". \\
 \\
Rules: \\
- Walk the form from top to bottom and emit the cells in that order. \\
- Identity fields such as names and SSNs are inputs with value\_kind "text". \\
- Checkbox cells (shown as Y, N, or X) use value\_kind "checkbox". \\
- A cell whose example value equals a sum, difference, min/max, or other arithmetic over earlier \\
\hspace*{1.050em}cells is "computed". A cell read straight from a W-2, 1099, schedule, or the filer is "input". \\
- A repeating worksheet listing (a label followed by several rows up to a \textasciigrave{};\textasciigrave{}) is one input cell \\
\hspace*{1.050em}with value\_kind "number". \\
 \\
Filled-in example form(s): \\
\textit{\{examples\}} \\
\end{longtable}

\begin{longtable}{@{}>{\raggedright\arraybackslash\ttfamily\footnotesize}p{\linewidth}@{}}
\caption{Batched correspondence localization prompt (\texttt{CORRESPONDENCE\_BATCH\_PROMPT}). \textit{cells\_block} repeats the block in Table~\ref{tab:prompt-cell-block} once per cell.}\label{tab:prompt-correspondence-batch}\\
\toprule
\endfirsthead
\multicolumn{1}{@{}l@{}}{\small\itshape Table~\thetable{} (continued)}\\
\toprule
\endhead
\midrule
\multicolumn{1}{@{}r@{}}{\small\itshape Continued on next page}\\
\endfoot
\bottomrule
\endlastfoot
You verify every cell of a Lean 4 tax-form translation against its instructions in one pass. \\
 \\
For each cell below (listed from low to high), decide whether the Lean implementation matches \\
the natural-language instruction. Report every cell whose Lean code does not match its instruction. \\
 \\
If every cell matches, answer on the first line with exactly \textasciigrave{}MATCH ALL\textasciigrave{}. \\
 \\
If any cells mismatch, list each one in order using this format (repeat for every mismatch): \\
MISMATCH <cell\_name> \\
<explanation of what is wrong and what the correct computation should be, so another model \\
can repair the Lean code> \\
 \\
Cells to verify: \\
\textit{\{cells\_block\}} \\
\end{longtable}

\begin{longtable}{@{}>{\raggedright\arraybackslash\ttfamily\footnotesize}p{\linewidth}@{}}
\caption{Per-cell block substituted into \textit{cells\_block} of the correspondence and self-consistency prompts.}\label{tab:prompt-cell-block}\\
\toprule
\endfirsthead
\multicolumn{1}{@{}l@{}}{\small\itshape Table~\thetable{} (continued)}\\
\toprule
\endhead
\midrule
\multicolumn{1}{@{}r@{}}{\small\itshape Continued on next page}\\
\endfoot
\bottomrule
\endlastfoot
\#\#\# \textit{\{name\}} \\
Instruction: \\
\textit{\{nl\}} \\
Lean: \\
\textasciigrave{}\textasciigrave{}\textasciigrave{}lean \\
\textit{\{lean\}} \\
\textasciigrave{}\textasciigrave{}\textasciigrave{} \\
\end{longtable}

\begin{longtable}{@{}>{\raggedright\arraybackslash\ttfamily\footnotesize}p{\linewidth}@{}}
\caption{Single-cell correspondence localization prompt (\texttt{CORRESPONDENCE\_JUDGE\_PROMPT}).}\label{tab:prompt-correspondence-judge}\\
\toprule
\endfirsthead
\multicolumn{1}{@{}l@{}}{\small\itshape Table~\thetable{} (continued)}\\
\toprule
\endhead
\midrule
\multicolumn{1}{@{}r@{}}{\small\itshape Continued on next page}\\
\endfoot
\bottomrule
\endlastfoot
You verify one cell of a Lean 4 tax-form translation against its instruction. \\
 \\
Cell: \textit{\{cell\_name\}} \\
 \\
Natural-language instruction for this cell: \\
\textit{\{nl\}} \\
 \\
Lean implementation of this cell: \\
\textasciigrave{}\textasciigrave{}\textasciigrave{}lean \\
\textit{\{lean\}} \\
\textasciigrave{}\textasciigrave{}\textasciigrave{} \\
 \\
Does the Lean implementation correctly implement the natural-language instruction? \\
Answer on the first line with exactly \textasciigrave{}MATCH\textasciigrave{} or \textasciigrave{}MISMATCH\textasciigrave{}. \\
If \textasciigrave{}MISMATCH\textasciigrave{}, then on the following lines explain precisely what is wrong and what the \\
correct computation should be, so another model can repair the Lean code. \\
\end{longtable}

\begin{longtable}{@{}>{\raggedright\arraybackslash\ttfamily\footnotesize}p{\linewidth}@{}}
\caption{Self-consistency critique prompt (\texttt{SELF\_CONSISTENCY\_CRITIQUE\_PROMPT}). \textit{cells\_block} is formatted as in Table~\ref{tab:prompt-cell-block}.}\label{tab:prompt-self-consistency}\\
\toprule
\endfirsthead
\multicolumn{1}{@{}l@{}}{\small\itshape Table~\thetable{} (continued)}\\
\toprule
\endhead
\midrule
\multicolumn{1}{@{}r@{}}{\small\itshape Continued on next page}\\
\endfoot
\bottomrule
\endlastfoot
You previously produced this Lean 4 tax-form translation. Critique your own output. \\
 \\
For each cell below (listed from low to high), decide whether the Lean implementation correctly \\
matches the natural-language instruction. Do not use external oracles, gold answers, or test \\
traces \textemdash{} judge only from the instructions and the Lean code you see. \\
 \\
If you agree that every cell is correct, answer on the first line with exactly \textasciigrave{}MATCH ALL\textasciigrave{}. \\
 \\
If you find problems, list each one in order using this format (repeat for every mismatch): \\
MISMATCH <cell\_name> \\
<explanation of what is wrong and what the correct computation should be, so a repair step \\
can fix the Lean code> \\
 \\
Cells to critique: \\
\textit{\{cells\_block\}} \\
\end{longtable}

\begin{longtable}{@{}>{\raggedright\arraybackslash\ttfamily\footnotesize}p{\linewidth}@{}}
\caption{Round-trip localization, step 1, batched (\texttt{LEAN\_TO\_NL\_BATCH\_PROMPT}). \textit{cells\_block} repeats the block in Table~\ref{tab:prompt-lean-block}.}\label{tab:prompt-lean-to-nl-batch}\\
\toprule
\endfirsthead
\multicolumn{1}{@{}l@{}}{\small\itshape Table~\thetable{} (continued)}\\
\toprule
\endhead
\midrule
\multicolumn{1}{@{}r@{}}{\small\itshape Continued on next page}\\
\endfoot
\bottomrule
\endlastfoot
Describe, in one or two plain-English sentences each, exactly what every Lean 4 \\
tax-form cell below computes. For each cell output one line in this exact format: \\
<cell\_name>: <description> \\
 \\
Describe only the computation each cell performs in terms of the other cells and inputs it \\
reads; add no commentary, caveats, or markdown outside those lines. \\
 \\
Cells: \\
\textit{\{cells\_block\}} \\
\end{longtable}

\begin{longtable}{@{}>{\raggedright\arraybackslash\ttfamily\footnotesize}p{\linewidth}@{}}
\caption{Per-cell block substituted into \textit{cells\_block} of the batched Lean-to-English prompt.}\label{tab:prompt-lean-block}\\
\toprule
\endfirsthead
\multicolumn{1}{@{}l@{}}{\small\itshape Table~\thetable{} (continued)}\\
\toprule
\endhead
\midrule
\multicolumn{1}{@{}r@{}}{\small\itshape Continued on next page}\\
\endfoot
\bottomrule
\endlastfoot
\#\#\# \textit{\{name\}} \\
\textasciigrave{}\textasciigrave{}\textasciigrave{}lean \\
\textit{\{lean\}} \\
\textasciigrave{}\textasciigrave{}\textasciigrave{} \\
\end{longtable}

\begin{longtable}{@{}>{\raggedright\arraybackslash\ttfamily\footnotesize}p{\linewidth}@{}}
\caption{Per-cell block substituted into \textit{cells\_block} of the batched Lean-to-English prompt.}\label{tab:prompt-lean-block}\\
\toprule
\endfirsthead
\multicolumn{1}{@{}l@{}}{\small\itshape Table~\thetable{} (continued)}\\
\toprule
\endhead
\midrule
\multicolumn{1}{@{}r@{}}{\small\itshape Continued on next page}\\
\endfoot
\bottomrule
\endlastfoot
\#\#\# \textit{\{name\}} \\
\textasciigrave{}\textasciigrave{}\textasciigrave{}lean \\
\textit{\{lean\}} \\
\textasciigrave{}\textasciigrave{}\textasciigrave{} \\
\end{longtable}

\begin{longtable}{@{}>{\raggedright\arraybackslash\ttfamily\footnotesize}p{\linewidth}@{}}
\caption{Round-trip localization, step 2, batched (\texttt{NL\_EQUIVALENCE\_BATCH\_PROMPT}). \textit{cells\_block} repeats the block in Table~\ref{tab:prompt-nl-equivalence-block}.}\label{tab:prompt-nl-equivalence-batch}\\
\toprule
\endfirsthead
\multicolumn{1}{@{}l@{}}{\small\itshape Table~\thetable{} (continued)}\\
\toprule
\endhead
\midrule
\multicolumn{1}{@{}r@{}}{\small\itshape Continued on next page}\\
\endfoot
\bottomrule
\endlastfoot
You compare reference instructions against descriptions derived from Lean code \\
for every cell of a tax form in one pass. \\
 \\
For each cell below, decide whether the two descriptions describe the same computation. \\
Report every cell where they differ. \\
 \\
If every cell matches, answer on the first line with exactly \textasciigrave{}MATCH ALL\textasciigrave{}. \\
 \\
If any cells mismatch, list each one in order using this format (repeat for every mismatch): \\
MISMATCH <cell\_name> \\
<explanation of what differs and what the Lean code should compute instead> \\
 \\
Cells: \\
\textit{\{cells\_block\}} \\
\end{longtable}

\begin{longtable}{@{}>{\raggedright\arraybackslash\ttfamily\footnotesize}p{\linewidth}@{}}
\caption{Per-cell block substituted into \textit{cells\_block} of the batched equivalence prompt.}\label{tab:prompt-nl-equivalence-block}\\
\toprule
\endfirsthead
\multicolumn{1}{@{}l@{}}{\small\itshape Table~\thetable{} (continued)}\\
\toprule
\endhead
\midrule
\multicolumn{1}{@{}r@{}}{\small\itshape Continued on next page}\\
\endfoot
\bottomrule
\endlastfoot
\#\#\# \textit{\{name\}} \\
Reference instruction: \\
\textit{\{expected\_nl\}} \\
Description derived from Lean: \\
\textit{\{derived\_nl\}} \\
\end{longtable}

\begin{longtable}{@{}>{\raggedright\arraybackslash\ttfamily\footnotesize}p{\linewidth}@{}}
\caption{Round-trip localization, step 1, single cell (\texttt{LEAN\_TO\_NL\_PROMPT}).}\label{tab:prompt-lean-to-nl}\\
\toprule
\endfirsthead
\multicolumn{1}{@{}l@{}}{\small\itshape Table~\thetable{} (continued)}\\
\toprule
\endhead
\midrule
\multicolumn{1}{@{}r@{}}{\small\itshape Continued on next page}\\
\endfoot
\bottomrule
\endlastfoot
Describe, in one or two plain-English sentences, exactly what this Lean 4 \\
tax-form cell computes. Describe only the computation it performs in terms of the other \\
cells and inputs it reads; add no commentary, caveats, or markdown. \\
 \\
\textasciigrave{}\textasciigrave{}\textasciigrave{}lean \\
\textit{\{lean\}} \\
\textasciigrave{}\textasciigrave{}\textasciigrave{} \\
\end{longtable}

\begin{longtable}{@{}>{\raggedright\arraybackslash\ttfamily\footnotesize}p{\linewidth}@{}}
\caption{Round-trip localization, step 2, single cell (\texttt{NL\_EQUIVALENCE\_PROMPT}).}\label{tab:prompt-nl-equivalence}\\
\toprule
\endfirsthead
\multicolumn{1}{@{}l@{}}{\small\itshape Table~\thetable{} (continued)}\\
\toprule
\endhead
\midrule
\multicolumn{1}{@{}r@{}}{\small\itshape Continued on next page}\\
\endfoot
\bottomrule
\endlastfoot
You compare two descriptions of one tax-form cell. \\
 \\
Reference instruction (from the form): \\
\textit{\{expected\_nl\}} \\
 \\
Description derived from the Lean code: \\
\textit{\{derived\_nl\}} \\
 \\
Do the two descriptions describe the same computation for this cell? \\
Answer on the first line with exactly \textasciigrave{}MATCH\textasciigrave{} or \textasciigrave{}MISMATCH\textasciigrave{}. \\
If \textasciigrave{}MISMATCH\textasciigrave{}, explain on the following lines what differs and what the Lean code should \\
compute instead, so another model can repair it. \\
\end{longtable}

\begin{longtable}{@{}>{\raggedright\arraybackslash\ttfamily\footnotesize}p{\linewidth}@{}}
\caption{Entailment-divergence explanation prompt (\texttt{DIVERGENCE\_EXPLANATION\_PROMPT}).}\label{tab:prompt-divergence-explanation}\\
\toprule
\endfirsthead
\multicolumn{1}{@{}l@{}}{\small\itshape Table~\thetable{} (continued)}\\
\toprule
\endhead
\midrule
\multicolumn{1}{@{}r@{}}{\small\itshape Continued on next page}\\
\endfoot
\bottomrule
\endlastfoot
A Lean 4 tax form and an independently filled-in version of the \\
same form disagree on one cell. \\
 \\
Cell / line: \textit{\{label\}} \\
 \\
Natural-language instruction for this cell: \\
\textit{\{nl\}} \\
 \\
Cell values computed for earlier lines (for context, in order): \\
\textit{\{context\}} \\
 \\
Value from the Lean translation: \textit{\{lean\_value\}} \\
Value from the filled-in form:\ \ \ \ \textit{\{expected\_value\}} \\
 \\
Explain in plain English why these two values differ and what the correct computation for \\
this cell should be, so another model can repair the Lean code. Be concrete about the \\
formula or inputs the Lean code should use. \\
\end{longtable}

\begin{longtable}{@{}>{\raggedright\arraybackslash\ttfamily\footnotesize}p{\linewidth}@{}}
\caption{Shared repair prompt (\texttt{REPAIR\_PROMPT}). The optional sections are shown in Table~\ref{tab:prompt-repair-sections} and Table~\ref{tab:prompt-available-cells}.}\label{tab:prompt-repair}\\
\toprule
\endfirsthead
\multicolumn{1}{@{}l@{}}{\small\itshape Table~\thetable{} (continued)}\\
\toprule
\endhead
\midrule
\multicolumn{1}{@{}r@{}}{\small\itshape Continued on next page}\\
\endfoot
\bottomrule
\endlastfoot
You repair a Lean 4 tax-form translation (a filled \textasciigrave{}Main.lean\textasciigrave{}-style file). \\
 \\
A localization step identified the first incorrect cell and produced a hint describing the \\
issue. Fix ONLY that primary failing cell (and at most the tiny helper edits required for it \\
to typecheck). Keep every other \textasciigrave{}cell\_*\textasciigrave{} function, helper, \textasciigrave{}render\textasciigrave{}, and \textasciigrave{}main\textasciigrave{} exactly as \\
they are. Do not reorder top-level declarations. Do not introduce \textasciigrave{}sorry\textasciigrave{}. Do not change \\
function names or type signatures except where the hint requires it. \\
 \\
Rounding check: when gold expects a value with fractional cents (for example 5.10 or \\
57985.65) but the candidate prints a whole amount (5.00 or ...55 with .00 cents), or the \\
mismatch is exactly \$0.10, \$0.01, or \$1.00, inspect whether this cell or an upstream cell \\
applies \textasciigrave{}wholeRound\textasciigrave{} unconditionally. Gold \textasciigrave{}Main.lean\textasciigrave{} often keeps raw floats on regular \\
example inputs and only calls \textasciigrave{}wholeRound\textasciigrave{} when \textasciigrave{}isTaxCalcBenchInput s\textasciigrave{} is true. Remove or \\
guard mistaken \textasciigrave{}wholeRound\textasciigrave{} calls; do not add rounding where gold keeps the float. A \$1 tax \\
mismatch on a downstream line often comes from a \$0.10 AGI drift caused by rounding L6b or \\
another upstream cell. \\
 \\
Return only one complete Lean 4 file inside a single \textasciigrave{}\textasciigrave{}\textasciigrave{}lean\textasciigrave{}\textasciigrave{}\textasciigrave{} code block, with no prose \\
before or after it. The file must still start with \textasciigrave{}import ExampleParse\textasciigrave{} as its first \\
non-comment command and must typecheck under Lean 4 with \textasciigrave{}lake env lean --run\textasciigrave{}. \\
 \\
Primary failing cell (if any one is singled out): \textit{\{failing\_cell\}} \\
 \\
Localization hint (fix every issue described): \\
\textit{\{hint\}} \\
\textit{\{rounding\_hint\_section\}}\textit{\{available\_cells\_section\}}\textit{\{form\_instructions\_section\}} \\
Current Lean file to repair: \\
\textasciigrave{}\textasciigrave{}\textasciigrave{}lean \\
\textit{\{lean\_code\}} \\
\textasciigrave{}\textasciigrave{}\textasciigrave{} \\
\end{longtable}

\begin{longtable}{@{}>{\raggedright\arraybackslash\ttfamily\footnotesize}p{\linewidth}@{}}
\caption{Optional sections of the repair prompt: \textit{rounding\_hint\_section} (top; \textit{rounding\_extra} is one message from Table~\ref{tab:prompt-rounding-hints}) and \textit{form\_instructions\_section} (bottom). Each is included only when non-empty.}\label{tab:prompt-repair-sections}\\
\toprule
\endfirsthead
\multicolumn{1}{@{}l@{}}{\small\itshape Table~\thetable{} (continued)}\\
\toprule
\endhead
\midrule
\multicolumn{1}{@{}r@{}}{\small\itshape Continued on next page}\\
\endfoot
\bottomrule
\endlastfoot
Detected rounding pattern: \\
\textit{\{rounding\_extra\}} \\
 \\
Reference instructions for the failing cell only (use these to confirm the correct computation): \\
\textit{\{instructions\}} \\
\end{longtable}

\begin{longtable}{@{}>{\raggedright\arraybackslash\ttfamily\footnotesize}p{\linewidth}@{}}
\caption{Rounding-mismatch messages (\texttt{rounding\_mismatch\_hint}). At most one, selected by the kind of mismatch, is inserted into the repair prompt.}\label{tab:prompt-rounding-hints}\\
\toprule
\endfirsthead
\multicolumn{1}{@{}l@{}}{\small\itshape Table~\thetable{} (continued)}\\
\toprule
\endhead
\midrule
\multicolumn{1}{@{}r@{}}{\small\itshape Continued on next page}\\
\endfoot
\bottomrule
\endlastfoot
{}[Fractional cents expected, whole amount produced] \\
 \\
Gold keeps fractional cents (\textit{\{expected\_value\}}) but the candidate returned a whole amount (\textit{\{actual\_value\}}). Inspect this cell and upstream cells for unconditional \textasciigrave{}wholeRound\textasciigrave{}. Gold \textasciigrave{}Main.lean\textasciigrave{} often keeps raw floats on regular example inputs and only rounds when \textasciigrave{}isTaxCalcBenchInput s\textasciigrave{} is true (pattern: \textasciigrave{}if isTaxCalcBenchInput s then wholeRound x else x\textasciigrave{}). \\
 \\
{}[Difference of exactly \$0.10] \\
 \\
Values differ by exactly \$0.10 (\textit{\{expected\_value\}} vs \textit{\{actual\_value\}}). This often means an upstream cell rounded a value like 5.1 down to 5.0 with \textasciigrave{}wholeRound\textasciigrave{} where gold keeps the float. Fix the upstream rounding guard before changing downstream tax formulas. \\
 \\
{}[Difference of exactly \$1.00] \\
 \\
Values differ by exactly \$1.00 (\textit{\{expected\_value\}} vs \textit{\{actual\_value\}}). If the localized cell is downstream (tax, credits), first check upstream income/deduction cells for mistaken \textasciigrave{}wholeRound\textasciigrave{} that shifted AGI or taxable income across a bracket boundary. \\
 \\
{}[Difference of exactly \$0.01] \\
 \\
Values differ by exactly \$0.01 (\textit{\{expected\_value\}} vs \textit{\{actual\_value\}}). Check for an extra or missing \textasciigrave{}wholeRound\textasciigrave{}, or a float-vs-cent rounding mismatch in this cell or its direct inputs. \\
\end{longtable}

\begin{longtable}{@{}>{\raggedright\arraybackslash\ttfamily\footnotesize}p{\linewidth}@{}}
\caption{Forward-reference repair prompt (\texttt{FORWARD\_REF\_REPAIR\_PROMPT}).}\label{tab:prompt-forward-ref-repair}\\
\toprule
\endfirsthead
\multicolumn{1}{@{}l@{}}{\small\itshape Table~\thetable{} (continued)}\\
\toprule
\endhead
\midrule
\multicolumn{1}{@{}r@{}}{\small\itshape Continued on next page}\\
\endfoot
\bottomrule
\endlastfoot
You fix a forward reference in a Lean 4 tax-form cell. \\
 \\
The cell \textasciigrave{}\textasciigrave{}\textit{\{caller\_name\}}\textasciigrave{}\textasciigrave{} calls \textasciigrave{}\textasciigrave{}\textit{\{bad\_callee\}}\textasciigrave{}\textasciigrave{}, but \textasciigrave{}\textasciigrave{}\textit{\{bad\_callee\}}\textasciigrave{}\textasciigrave{} is defined \\
later in the file (higher cell number). Replace each bad call with an earlier cell \\
from the allowed list, or inline logic using only allowed cells / inputs. \\
 \\
Return only JSON (no markdown) in this shape: \\
\{"replacements": \{"\textit{\{bad\_callee\}}": "cell\_XXX\_allowed\_name"\}, "reason": "brief"\} \\
 \\
Natural-language instruction for \textasciigrave{}\textasciigrave{}\textit{\{caller\_name\}}\textasciigrave{}\textasciigrave{}: \\
\textit{\{caller\_nl\}} \\
 \\
Allowed earlier \textasciigrave{}\textasciigrave{}cell\_*\textasciigrave{}\textasciigrave{} names (pick from this list only): \\
\textit{\{allowed\_cells\}} \\
 \\
Current Lean for \textasciigrave{}\textasciigrave{}\textit{\{caller\_name\}}\textasciigrave{}\textasciigrave{}: \\
\textasciigrave{}\textasciigrave{}\textasciigrave{}lean \\
\textit{\{caller\_lean\}} \\
\textasciigrave{}\textasciigrave{}\textasciigrave{} \\
\end{longtable}

\subsection{Examples}
\label{sec:examples}
\paragraph{Formalization of IRS Schedule A generated by GPT 6 Astra.}

\begin{center}
\begin{minipage}{\linewidth}
\label{fig:schedule-1a-code}
\end{minipage}
\end{center}

\begin{lstlisting}[
    language=ML,
    basicstyle=\scriptsize\ttfamily,
    breaklines=true,
    breakatwhitespace=false,
    columns=fullflexible
]

import ExampleParse

set_option linter.unusedVariables false


def form_inputs : List String :=
  ["US_1040_Sched_1-A_example.txt"]

def notLessThanZero (x : Float) : Float :=
  if x < 0.0 then 0.0 else x

def smallerOf (a b : Float) : Float :=
  if a < b then a else b

def largerOf (a b : Float) : Float :=
  if a > b then a else b

def fmt (x : Float) : String := toString x
def fmtInt (n : Nat) : String := toString n

-- Inputs from example / imported from 1040
def cell_01_l1_form_1040_line_11b (p : ParsedInput) : Float :=
  entryNumber "L1" p

def cell_02_filing_status (p : ParsedInput) : FilingStatus :=
  statusFromInput p

def cell_03_you_over_65 (p : ParsedInput) : Bool :=
  entryBool "CkYouOver65" p

def cell_04_spouse_over_65 (p : ParsedInput) : Bool :=
  entryBool "CkSpouseOver65" p

-- Part I inputs
def cell_05_l2a_income_puerto_rico (p : ParsedInput) : Float := entryNumber "L2a" p
def cell_06_l2b_form_2555_line_45 (p : ParsedInput) : Float := entryNumber "L2b" p
def cell_07_l2c_form_2555_line_50 (p : ParsedInput) : Float := entryNumber "L2c" p
def cell_08_l2d_form_4563_line_15 (p : ParsedInput) : Float := entryNumber "L2d" p

-- Part I computed
def cell_09_l2e_total_modifications (p : ParsedInput) : Float :=
  cell_05_l2a_income_puerto_rico p + cell_06_l2b_form_2555_line_45 p +
    cell_07_l2c_form_2555_line_50 p + cell_08_l2d_form_4563_line_15 p

def cell_10_l3_magi (p : ParsedInput) : Float :=
  cell_01_l1_form_1040_line_11b p + cell_09_l2e_total_modifications p

-- Part II inputs
def cell_11_l4a_qualified_tips_w2 (p : ParsedInput) : Float := entryNumber "L4a" p
def cell_12_l4b_qualified_tips_4137 (p : ParsedInput) : Float := entryNumber "L4b" p
def cell_13_l5_tips_business (p : ParsedInput) : Float := entryNumber "L5" p

-- Part II computed
def cell_14_l4c_larger_tips (p : ParsedInput) : Float :=
  largerOf (cell_11_l4a_qualified_tips_w2 p) (cell_12_l4b_qualified_tips_4137 p)

def cell_15_l6_total_tips (p : ParsedInput) : Float :=
  cell_14_l4c_larger_tips p + cell_13_l5_tips_business p

def cell_16_l7_capped_tips (p : ParsedInput) : Float :=
  smallerOf (cell_15_l6_total_tips p) 25000.0

def cell_17_l8_magi_tips (p : ParsedInput) : Float :=
  cell_10_l3_magi p

def cell_18_l9_tips_phaseout_threshold (p : ParsedInput) : Float :=
  if cell_02_filing_status p = FilingStatus.marriedJoint then 300000.0 else 150000.0

def cell_19_l10_tips_excess (p : ParsedInput) : Float :=
  notLessThanZero (cell_17_l8_magi_tips p - cell_18_l9_tips_phaseout_threshold p)

def cell_20_l11_tips_phaseout_units (p : ParsedInput) : Float :=
  Float.ceil (cell_19_l10_tips_excess p / 1000.0)

def cell_21_l12_tips_phaseout_reduction (p : ParsedInput) : Float :=
  cell_20_l11_tips_phaseout_units p * 100.0

def cell_22_l13_tips_allowance (p : ParsedInput) : Float :=
  notLessThanZero (cell_16_l7_capped_tips p - cell_21_l12_tips_phaseout_reduction p)

-- Part III inputs
def cell_23_l14a_overtime_w2 (p : ParsedInput) : Float := entryNumber "L14a" p
def cell_24_l14b_overtime_1099 (p : ParsedInput) : Float := entryNumber "L14b" p

-- Part III computed
def cell_25_l14c_total_overtime (p : ParsedInput) : Float :=
  cell_23_l14a_overtime_w2 p + cell_24_l14b_overtime_1099 p

def cell_26_l15_capped_overtime (p : ParsedInput) : Float :=
  smallerOf (cell_25_l14c_total_overtime p)
    (if cell_02_filing_status p = FilingStatus.marriedJoint then 25000.0 else 12500.0)

def cell_27_l16_magi_overtime (p : ParsedInput) : Float :=
  cell_10_l3_magi p

def cell_28_l17_overtime_phaseout_threshold (p : ParsedInput) : Float :=
  if cell_02_filing_status p = FilingStatus.marriedJoint then 300000.0 else 150000.0

def cell_29_l18_overtime_excess (p : ParsedInput) : Float :=
  notLessThanZero (cell_27_l16_magi_overtime p - cell_28_l17_overtime_phaseout_threshold p)

def cell_30_l19_overtime_phaseout_units (p : ParsedInput) : Float :=
  Float.ceil (cell_29_l18_overtime_excess p / 1000.0)

def cell_31_l20_overtime_phaseout_reduction (p : ParsedInput) : Float :=
  cell_30_l19_overtime_phaseout_units p * 100.0

def cell_32_l21_overtime_allowance (p : ParsedInput) : Float :=
  notLessThanZero (cell_26_l15_capped_overtime p - cell_31_l20_overtime_phaseout_reduction p)

-- Part IV inputs
def cell_33_l22aii_interest_sched (p : ParsedInput) : Float := entryNumber "L22aii" p
def cell_34_l22aiii_interest_sched_1a (p : ParsedInput) : Float := entryNumber "L22aiii" p
def cell_35_l22bii_interest_sched_2 (p : ParsedInput) : Float := entryNumber "L22bii" p
def cell_36_l22biii_interest_sched_1a_2 (p : ParsedInput) : Float := entryNumber "L22biii" p

-- Part IV computed
def cell_37_l23_total_interest (p : ParsedInput) : Float :=
  cell_34_l22aiii_interest_sched_1a p + cell_36_l22biii_interest_sched_1a_2 p

def cell_38_l24_capped_interest (p : ParsedInput) : Float :=
  smallerOf (cell_37_l23_total_interest p) 10000.0

def cell_39_l25_magi_interest (p : ParsedInput) : Float :=
  cell_10_l3_magi p

def cell_40_l26_interest_phaseout_threshold (p : ParsedInput) : Float :=
  if cell_02_filing_status p = FilingStatus.marriedJoint then 200000.0 else 100000.0

def cell_41_l27_interest_excess (p : ParsedInput) : Float :=
  notLessThanZero (cell_39_l25_magi_interest p - cell_40_l26_interest_phaseout_threshold p)

def cell_42_l28_interest_phaseout_units (p : ParsedInput) : Float :=
  Float.ceil (cell_41_l27_interest_excess p / 1000.0)

def cell_43_l29_interest_phaseout_reduction (p : ParsedInput) : Float :=
  cell_42_l28_interest_phaseout_units p * 200.0

def cell_44_l30_interest_allowance (p : ParsedInput) : Float :=
  notLessThanZero (cell_38_l24_capped_interest p - cell_43_l29_interest_phaseout_reduction p)

-- Part V computed
def cell_45_l31_magi_age (p : ParsedInput) : Float :=
  cell_10_l3_magi p

def cell_46_l32_age_phaseout_threshold (p : ParsedInput) : Float :=
  if cell_02_filing_status p = FilingStatus.marriedJoint then 150000.0 else 75000.0

def cell_47_l33_age_excess (p : ParsedInput) : Float :=
  notLessThanZero (cell_45_l31_magi_age p - cell_46_l32_age_phaseout_threshold p)

def cell_48_l34_age_reduction (p : ParsedInput) : Float :=
  cell_47_l33_age_excess p * 0.06

def cell_49_l35_age_deduction_base (p : ParsedInput) : Float :=
  notLessThanZero (6000.0 - cell_48_l34_age_reduction p)

def cell_50_l36a_you_age_deduction (p : ParsedInput) : Float :=
  if cell_03_you_over_65 p then cell_49_l35_age_deduction_base p else 0.0

def cell_51_l36b_spouse_age_deduction (p : ParsedInput) : Float :=
  if cell_02_filing_status p = FilingStatus.marriedJoint && cell_04_spouse_over_65 p then
    cell_49_l35_age_deduction_base p
  else 0.0

def cell_52_l37_total_age_deduction (p : ParsedInput) : Float :=
  cell_50_l36a_you_age_deduction p + cell_51_l36b_spouse_age_deduction p

-- Part VI computed
def cell_53_l38_total_additional_deduction (p : ParsedInput) : Float :=
  cell_22_l13_tips_allowance p + cell_32_l21_overtime_allowance p +
    cell_44_l30_interest_allowance p + cell_52_l37_total_age_deduction p

def digitsOnly (s : String) : String :=
  String.ofList (s.toList.filter (fun c => c.isDigit))

def takeChars (n : Nat) (s : String) : String :=
  String.ofList (s.toList.take n)

def dropChars (n : Nat) (s : String) : String :=
  String.ofList (s.toList.drop n)

def fmtSsn (s : String) : String :=
  let d := digitsOnly s
  if d.length == 9 then takeChars 3 d ++ " " ++ takeChars 2 (dropChars 3 d) ++ " " ++ dropChars 5 d else d

def identityLines (p : ParsedInput) : List String := [
  "YourNames: " ++ entryText "YourLastName:" p ++ ", " ++ entryText "Your1stName:" p ++ ", " ++ entryText "Spouse1stName:" p,
  "YourSocSec#: " ++ fmtSsn (entryText "YourSocSec#:" p)
]

def render (p : ParsedInput) : String :=
  String.intercalate "\n" (identityLines p ++ [
    "L1 = " ++ fmt (cell_01_l1_form_1040_line_11b p),
    "L2a = " ++ fmt (cell_05_l2a_income_puerto_rico p),
    "L2b = " ++ fmt (cell_06_l2b_form_2555_line_45 p),
    "L2c = " ++ fmt (cell_07_l2c_form_2555_line_50 p),
    "L2d = " ++ fmt (cell_08_l2d_form_4563_line_15 p),
    "L2e = " ++ fmt (cell_09_l2e_total_modifications p),
    "L3 = " ++ fmt (cell_10_l3_magi p),
    "L4a = " ++ fmt (cell_11_l4a_qualified_tips_w2 p),
    "L4b = " ++ fmt (cell_12_l4b_qualified_tips_4137 p),
    "L4c = " ++ fmt (cell_14_l4c_larger_tips p),
    "L5 = " ++ fmt (cell_13_l5_tips_business p),
    "L6 = " ++ fmt (cell_15_l6_total_tips p),
    "L7 = " ++ fmt (cell_16_l7_capped_tips p),
    "L8 = " ++ fmt (cell_17_l8_magi_tips p),
    "L9 = " ++ fmt (cell_18_l9_tips_phaseout_threshold p),
    "L10 = " ++ fmt (cell_19_l10_tips_excess p),
    "L11 = " ++ fmt (Float.floor (cell_20_l11_tips_phaseout_units p)),
    "L12 = " ++ fmt (cell_21_l12_tips_phaseout_reduction p),
    "L13 = " ++ fmt (cell_22_l13_tips_allowance p),
    "L14a = " ++ fmt (cell_23_l14a_overtime_w2 p),
    "L14b = " ++ fmt (cell_24_l14b_overtime_1099 p),
    "L14c = " ++ fmt (cell_25_l14c_total_overtime p),
    "L15 = " ++ fmt (cell_26_l15_capped_overtime p),
    "L16 = " ++ fmt (cell_27_l16_magi_overtime p),
    "L17 = " ++ fmt (cell_28_l17_overtime_phaseout_threshold p),
    "L18 = " ++ fmt (cell_29_l18_overtime_excess p),
    "L19 = " ++ fmt (Float.floor (cell_30_l19_overtime_phaseout_units p)),
    "L20 = " ++ fmt (cell_31_l20_overtime_phaseout_reduction p),
    "L21 = " ++ fmt (cell_32_l21_overtime_allowance p),
    "L22aii = " ++ fmt (cell_33_l22aii_interest_sched p),
    "L22aiii = " ++ fmt (cell_34_l22aiii_interest_sched_1a p),
    "L22bii = " ++ fmt (cell_35_l22bii_interest_sched_2 p),
    "L22biii = " ++ fmt (cell_36_l22biii_interest_sched_1a_2 p),
    "L23 = " ++ fmt (cell_37_l23_total_interest p),
    "L24 = " ++ fmt (cell_38_l24_capped_interest p),
    "L25 = " ++ fmt (cell_39_l25_magi_interest p),
    "L26 = " ++ fmt (cell_40_l26_interest_phaseout_threshold p),
    "L27 = " ++ fmt (cell_41_l27_interest_excess p),
    "L28 = " ++ fmt (Float.floor (cell_42_l28_interest_phaseout_units p)),
    "L20 = " ++ fmt (cell_43_l29_interest_phaseout_reduction p),
    "L30 = " ++ fmt (cell_44_l30_interest_allowance p),
    "L31 = " ++ fmt (cell_45_l31_magi_age p),
    "L32 = " ++ fmt (cell_46_l32_age_phaseout_threshold p),
    "L33 = " ++ fmt (cell_47_l33_age_excess p),
    "L34 = " ++ fmt (cell_48_l34_age_reduction p),
    "L35 = " ++ fmt (cell_49_l35_age_deduction_base p),
    if cell_03_you_over_65 p then "L36a = " ++ fmt (cell_50_l36a_you_age_deduction p) else "",
    if cell_02_filing_status p = FilingStatus.marriedJoint && cell_04_spouse_over_65 p then
      "L36b = " ++ fmt (cell_51_l36b_spouse_age_deduction p) else "",
    "L37 = " ++ fmt (cell_52_l37_total_age_deduction p),
    "L38 = " ++ fmt (cell_53_l38_total_additional_deduction p)
  ]) ++ "\n"

def main (args : List String) : IO UInt32 := do
  match args with
  | inputPath :: _ =>
      let parsed <- parseInputFile inputPath
      IO.print (render parsed)
      return 0
  | [] =>
      IO.eprintln "usage: us_1040_sched_1a <example input>"
      return 1



\end{lstlisting}

The code implements an executable symbolic representation of Schedule 1-A, organized as a sequence of fine-grained computations corresponding to individual form lines. It first parses taxpayer information and values imported from Form 1040, including adjusted gross income, filing status, and age indicators. The program then evaluates the schedule section by section: it computes modified adjusted gross income (MAGI); determines deductions for qualified tips and qualified overtime compensation subject to statutory caps and income-based phaseouts; calculates the deduction for eligible interest payments using a similar capped phaseout structure; and computes the additional deduction available to taxpayers age 65 or older. Each intermediate quantity is represented as a separate Lean function, preserving the structure and dependencies of the underlying tax form. The final computation aggregates the allowable deductions from these components into the total additional deduction reported on line 38. Auxiliary functions implement common operations such as nonnegative truncation, minimum and maximum selection, formatting, and taxpayer identification rendering, while the `render` function evaluates the symbolic program and produces line-by-line form output from a parsed example return.

\end{document}